\documentclass[iicol,sn-apa]{sn-jnl}

\usepackage{mathrsfs}
\usepackage{float}

\usepackage{amsmath,amssymb,amsfonts}
\usepackage{amsthm}
\usepackage{mathtools}
\usepackage{mathrsfs}
\usepackage{siunitx}

\usepackage{graphicx}
\usepackage{array}
\usepackage{multirow}
\usepackage{booktabs}
\usepackage{rotating}  

\usepackage{algorithm}
\usepackage{algorithmicx}
\usepackage{algpseudocode}

\usepackage{xcolor}
\usepackage[dvipsnames]{xcolor}
\usepackage{textcomp}
\usepackage{listings}

\usepackage{comment}
\usepackage{manyfoot}
\usepackage[title]{appendix}

\usepackage[capitalize]{cleveref}
\crefname{section}{Sec.}{Secs.}
\crefname{section}{Section}{Sections}
\crefname{table}{Table}{Tables}
\crefname{table}{Tab.}{Tabs.}

\usepackage{hyperref}

\begin{document}

\title[Article Title]{Robust Shape from Focus via Multiscale Directional Dilated Laplacian and Recurrent Network}

\author[1]{\fnm{Khurram} \sur{Ashfaq}}\email{khurram@koreatech.ac.kr }

\author*[1]{\fnm{Muhammad Tariq} \sur{Mahmood}}\email{tariq@koreatech.ac.kr}

\affil[1]{\orgdiv{Future Convergence Engineering, School of Computer Science and Engineering}, \orgname{Korea University of Technology and Education}, \orgaddress{\city{Cheonan}, \postcode{31253},  \country{Republic of Korea}}}

\abstract{ Shape-from-Focus (SFF) is a passive depth estimation technique that infers scene depth by analyzing focus variations in a focal stack. Most recent deep learning–based SFF methods typically operate in two stages: first, they extract focus volumes (a per pixel representation of focus likelihood across the focal stack) using heavy feature encoders; then, they estimate depth via a simple one-step aggregation technique that often introduces artifacts and amplifies noise in the depth map. To address these issues, we propose a hybrid framework. Our method computes multi-scale focus volumes traditionally using handcrafted Directional Dilated Laplacian (DDL) kernels, which capture long-range and directional focus variations to form robust focus volumes. These focus volumes are then fed into a lightweight, multi-scale GRU-based depth extraction module that iteratively refines an initial depth estimate at a lower resolution for computational efficiency. Finally, a learned convex upsampling module within our recurrent network reconstructs high-resolution depth maps while preserving fine scene details and sharp boundaries. Extensive experiments on both synthetic and real-world datasets demonstrate that our approach outperforms state-of-the-art deep learning and traditional methods, achieving superior accuracy and generalization across diverse focal conditions. Our code is available at \href{https://github.com/khurramashfaq/ddl-gru-sff}{https://github.com/khurramashfaq/ddl-gru-sff}.}


\maketitle

\section{Introduction}\label{sec1}

Shape-from-focus (SFF), also known as depth-from-focus, is a passive depth estimation technique that exploits variations in image sharpness across a focal stack, an ordered sequence of images captured at different focal settings, to infer scene depth. In this approach, each scene point reaches its maximum sharpness at a specific focal distance, allowing depth to be estimated by identifying the focal plane at which the point is optimally focused \citep{yan2025sas}. It is widely applied in 3D measurement, providing accurate spatial analysis for industrial inspection, precision manufacturing, virtual and augmented reality (VR/AR), autonomous navigation, and archaeological studies \citep{zheng2025depth}.

In general, SFF methods can be broadly categorized into traditional and deep learning–based approaches. The traditional process involves evaluating the focus quality of each pixel in every image of the focal stack by applying a predefined operator, known as a focus measure (FM). This operation produces a focus volume (FV) that mirrors the dimensions of the image sequence, offering three-dimensional focus information for every pixel. A depth map is then extracted from the FV based on the image index corresponding to the most focused frame for each pixel. However, these methods are often susceptible to noise, depend critically on the choice of focus metrics, and may struggle to capture gradual depth variations \citep{ashfaq2026dual}.

Deep learning–based SFF methods, on the other hand, harness the power of neural networks to predict depth. Early approaches \citep{hazirbas2019deep} attempted to directly derive depth using end-to-end models by creating deep feature volumes and then using a $1 \times 1$ convolution for channel reduction to regress depth; however, these efforts did not yield satisfactory results due to the aggressive channel-wise reduction discarding important spatial and cross-channel context, leading to ambiguous depth estimates. To address these shortcomings, recent methods have adopted a strategy inspired by traditional techniques: they first construct a three-dimensional focus volume that encodes per-pixel focus likelihood across the focal stack and then extract depth from this volume. This approach has introduced two major challenges. First, the use of heavy deep encoders with an extensive number of parameters. Second, instead of employing a dedicated depth extraction module, these techniques typically collapse the three-dimensional focus volume into a two-dimensional depth map in a single step through straightforward aggregation techniques. For example, \citep{wang2021bridging} and \citep{won2022learning} apply a softplus function to smooth the focus volume before normalizing it along the focal dimension to obtain focus probabilities, and then extract depth by identifying the focal slice with the highest focus probability. In contrast, \citep{yang2022deep} directly applies softmax along the focal dimension and then extract depth, while \citep{fujimura2024deep} uses softmax followed by a soft argmin operator for depth extraction. This one-step reduction can amplify noise, discard important spatial context, and lead to inaccurate depth predictions, particularly around object boundaries and in regions with subtle focus transitions. As a result, the final depth maps often exhibit unwanted artifacts and blurred discontinuities.

This paper proposes a hybrid approach for SFF that computes the focus volume using a traditional method and subsequently extracts depth through a Gated Recurrent Units (GRU)–based recurrent network. Initially, multi-scale focus volumes are obtained by convolving directional dilated Laplacian (DDL) kernels at varying dilation rates. These handcrafted kernels are robust to noise, capture long-range dependencies, and account for directional focus variations, thereby forming robust focus volumes. Since these focus volumes are computed outside the learning framework via simple convolution operations, the process is extremely fast. Subsequently, the focus volumes, along with some additional context features, are fed into a compact, multi-scale GRU-based depth extractor that treats depth estimation as a progressive refinement process. At each iteration, the GRU module updates the depth prediction at a lower resolution by incorporating both local focus cues and global contextual information. Furthermore, to enhance the resolution of the depth maps, a learned convex upsampling strategy is introduced to reconstruct fine details that might be lost during downsampling. By supervising intermediate depth predictions and progressively refining them, our method yields high-fidelity depth maps. Experimental results on both synthetic and real-world datasets demonstrate that our approach significantly enhances depth accuracy and spatial coherence relative to state-of-the-art methods. Moreover, our model exhibits superior generalization capabilities, achieving optimal performance when evaluated on real-world data.

The key contributions of this work can be summarized as: 
\begin{enumerate} 
\item \textbf{Directional Dilated Laplacian Kernels for Enhanced Focus Volumes:} We introduce multi-scale directional dilated Laplacian (DDL) kernels that are robust to noise, capture long-range dependencies, and account for directional focus variations, thereby forming robust focus volumes. 
\item \textbf{Deep Recurrent Model for Depth Extraction from Focus Volumes:} Rather than directly extracting depth from focus volumes using deep features derived from the focal stack, we propose a deep recurrent model that leverages our traditional focus volumes for progressive depth refinement. 
\item \textbf{Enhanced Depth Maps Compared to State-of-the-Art Models:} Our framework is lightweight and has been evaluated across diverse focal stacks, consistently generating high-quality depth maps that generalize well to unseen scenes, either outperforming or at least matching the performance of state-of-the-art models.
\end{enumerate}

\section{Background}\label{sec2}

\subsection{Depth Estimation}

Depth estimation is the process of determining the distance between objects and a camera, enabling machines to perceive and interact with their environment in three dimensions, a capability that is fundamental to computer vision, robotics, autonomous navigation, and 3D scene understanding. This technology finds diverse applications across multiple domains. For instance, in underwater exploration, depth estimation plays a pivotal role in image restoration by addressing the unique challenges posed by aquatic environments \citep{chang2018single}. In robotics, precise depth maps are critical for navigation and real-time interaction with surroundings \citep{dong2022towards}. Furthermore, in low-light conditions, enhancing depth information is essential for maintaining the performance of vision-based systems, as demonstrated by approaches that leverage depth cues to restore images captured under poor illumination \citep{wang2024multimodal}.

Techniques for depth estimation are broadly categorized into active and passive methods. Active methods \citep{rodrigues2020active} project known patterns or signals into the environment and analyze their reflections to compute depth; common approaches include structured light systems \citep{scharstein2003high}, which project a grid or dot pattern onto surfaces, and time-of-flight sensors \citep{gokturk2004time,riegler2019connecting} that measure the time taken for light to return to the sensor. In contrast, passive methods infer depth by analyzing visual information from images without emitting additional signals. One widely used passive approach is stereo matching \citep{hirschmuller2007evaluation}, which identifies corresponding points in two images taken from different viewpoints and triangulates depth from the resulting disparities. A comprehensive survey further explores deep stereo matching techniques \citep{tosi2025survey}, offering insights into recent advances in this domain. Following these developments, structure-from-motion techniques \citep{schonberger2016structure} deduce depth by observing the movement of objects across multiple frames, providing an alternative method for reconstructing 3D scene geometry. Recently, monocular depth estimation techniques that rely on single images have also gained prominence, yielding impressive results. The MiDaS framework \citep{ranftl2020towards} introduced a unified training strategy that combines multiple diverse datasets using scale- and shift-invariant loss functions. This approach effectively addresses the challenge of inconsistent depth annotations across datasets, producing models that generalize well to unseen scenes and exhibit strong cross-domain robustness. Building upon such progress, the more recent transformer-based Depth Anything V2 \citep{yang2024depth} leverages extensive pretraining on large and diverse image–depth pairs, achieving highly consistent performance across real-world conditions. Despite their success, these monocular approaches typically require massive amounts of training data and significant computational resources, which can limit their applicability in data-scarce or lightweight deployment scenarios. Despite these advances, many of the aforementioned depth estimation methods come with significant practical limitations. Approaches that require multiple cameras, meticulous calibration, dedicated sensors, or tightly controlled lighting setups can be impractical for lightweight, affordable, or portable systems.

\subsection{Shape from Focus}\label{sec:SFF}

\begin{figure*}[ht]
	\centering
	\includegraphics[width=\textwidth,keepaspectratio]{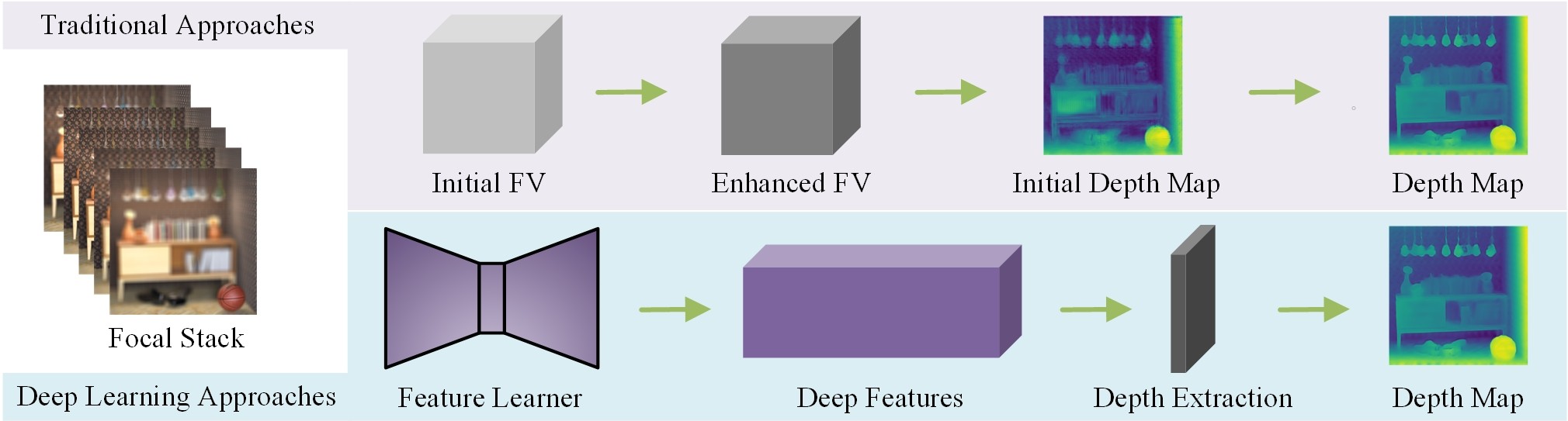}
	\caption{Pipelines for traditional and deep learning approaches in shape from focus systems.  To both approaches, a focal stack, comprising of multiple partially-focused images for the scene is provided as input. }
	\label{fig:Fig01}
\end{figure*}

Shape-from-focus (SFF) is a technique that estimates depth by analyzing the sharpness of objects across multiple images of a scene captured at different focus distances. Initially, a series of images is captured by varying the focus settings of the camera. Due to the shallow depth of field inherent in the camera, each image is only partially in focus. These images are then stacked together to form what is known as a focal stack.  Let the input focal stack be denoted by $ \mathbf{I} \in \mathbb{R}^{S\times 3 \times H \times W}$ consisting of $S$ images, each of having size $H \times W$, such that $\mathbf{I}^{(c)}_s(p)$ represents the image intensity at a pixel location $p$ for a color channel $c\in\{R,G,B\}$. Here, $p=(x,y)\in \mathbb{R}^2$ represents a pixel location, and $s\in\mathbb{R}$ denotes the image (frame) number in total images $S$. According to the thin lens law, there is a one-to-one correspondence between the maximally focused image patch and the depth map of that pixel. Therefore, in SFF, accurate measurement of the focus level or sharpness of the pixels is of paramount importance.    

SFF approaches can be divided into two broad categories: 1)  Traditional approaches, 2) Deep learning based Approaches.  The main steps involving in depth estimation through both approaches have been depicted in Fig.~\ref{fig:Fig01}.

\subsubsection{Traditional Approaches}\label{sec:CA}

In traditional approaches, first, the focus level computation for each pixel of the images in the focal stack is carried out by applying a focus measure $O(.)$ operator on the focal stack $\mathbf{I}^{(c)}_s(p)$. The focus measure operation provides the quantitative measure of sharpness for each pixel and results in an initial FV. This initial FV is 3D volumetric data having the same spatial dimensions as the input focal stack, except that it typically consists of a single channel instead of one for each color. Mathematically, it can be expressed as:

\begin{equation}
	\mathbf{I}^{\prime(c)}_s(p) = O(\mathbf{I}^{(c)}_s(p))
\end{equation}
In the literature, a variety of focus measures have been proposed. Previous works on FMs can be categorized into derivative-based, transform-based, statistics-based, and miscellaneous methods~\citep{pertuz2013analysis}. These methods leverage the property that focused images are sharper than defocused ones. Derivative-based methods usually use the gradient and second-order derivative to evaluate the image sharpness. By applying an FM,a  focus score is obtained for each image pixel of the focal stack. In~\citep{hurtado2018focus}, the modulus of the gradient of the color channels is utilized for focus measurements for auto-focusing process and extending the depth of field. A robust FM based on a ring and disk-like mask has also been proposed to tackle the noise~\citep{jeon2019ring}.  Due to the limited capabilities of FMs, the focus values in the initial FV are inaccurate. Therefore, at the next step, focus values are improved by applying an improvement scheme $E(.)$, and it results in a refined focus volume $\mathbf{I}^{\prime\prime(c)}_s(p)$, expressed a:
\begin{equation}
	\mathbf{I}^{\prime\prime(c)}_s(p) =E(\mathbf{I}^{\prime(c)}_s(p)).
\end{equation}

Among various methods for enhancement of initial focus volume, one simple and straightforward approach is the aggregation of focus values in a local neighborhood, i.e., by finding the average in window of different sizes around a central pixel \citep{nayar1994shape, thelen2009improvements}. Generally, smaller window sizes are not sufficient to overcome the noisy measurements, while the larger window sizes degrade the shape by over-smoothing it \citep{malik2007consideration}. In addition, irrespective of the window size, focus values get smoothed uniformly in linear filtering without considering the local structure or shape of the object. Consequently, to overcome these inherent drawbacks of linear filtering, researchers have considered the nonlinear filtering. Though computationally expensive, nonlinear filtering yields improved depth maps while retaining edges. In  \citep{aydin2010occlusion}, authors  proposed to compute weights and shape of the neighborhood window adaptively from the all-focused image. Similarly,  In \citep{mahmood2012nonlinear},  anisotropic diffusion filtering is applied to improve the FV  in which weights have been computed from the local structures. Further, optimization-based methods have been also applied to enhance the FV. For instance, in \citep{ahmad2005heuristic}  dynamic programming  is applied to optimize optimizing the focus measure in 3D volume. In \citep{shim2010novel} presented the SFF as a combinatorial optimization problem. The initial depth map is updated from the intermediate image volume which is generated from the local neighborhood. In \citep{tseng2014shape}, a MAP estimation solution is suggested with the inclusion of a spatially consistent model to better recover the depth over low contrast regions. In \citep{kumar2017accurate}, low rank prior as a regularizer is utilized, in the form of weighted nuclear norm minimization, to recover the 3D shape. In \citep{ali2021robust}, the problem of refining FV is formulated as an energy minimization framework that employs a nonconvex regularizer and incorporates shape priors. Nevertheless, conventional approaches still process pixels independently without fully exploiting sequence-level associations, whereas the recently proposed framework \citep{yan2025sas} addresses this limitation by treating the entire focal stack as complete 3D data. In addition, a dual-stage focus measure has been introduced to further improve accuracy by effectively leveraging information from all color channels \citep{ashfaq2026dual}.

The next step is to determine the focus score frame with the optimal focal score for each pixel location independently. In conventional approaches, a simple strategy of Winner-Takes-All is applied. According to this simple strategy, the focal indices of the images having the maximum focus values along the optical axis are taken as the depth value.
\begin{align}
\hat{D}(p) &= argmax_s (\mathbf{I}^{\prime\prime(c)}_s(p)), \label{depth_trad} \\
\mathbf{I}^{(c)}_{aif}(p) &= \mathbf{I}^{(c)}_{\hat{D}}(p),
\end{align}
where $\mathbf{I}^{(c)}_{aif}(p)$ denotes the all-in-focus image that can be obtained through focal stack once a depth map is available. Finally, the initial $\hat{D}(p)$ is further improved by applying a post-processing step. For instance, In \citep{moeller2015variational} a variational framework is implemented using total variation regularization to improve the initial depth map. In \citep{gaganov2009robust}, Markov Random Field energy is optimized for better depth maps.

\subsubsection{Deep Learning Based Approaches}\label{sec:DLA}

Deep learning-based SFF methods utilize neural networks to predict depth directly from the focal stack. Notably, these approaches have demonstrated robust generalization while requiring significantly fewer training scenes compared to monocular depth estimation techniques; for example, models trained on the FlyingThings3D \citep{mayer2016large} dataset, which comprises 1000 scenes, have shown strong performance on unseen data. This efficiency is largely attributed to the multiple observations of the same scene under different focus conditions, which provide a rich depth signal. These approaches first compute deep features from the input focal stack by applying convolutional neural networks (CNNs) and then predict depth maps from the deep features by projecting the volumetric data on 2D space \citep{wang2021bridging}. For deep feature extraction, in literature,  various architectures, including 2D encoder-decoder (2DED) \citep{hazirbas2019deep} and 3D encoder-decoder (3DED) networks \citep{wang2021bridging} have been utilized. In 2DED modules, 2D convolutions are applied to each image in the stack individually, and feature maps from the individual images are concatenated to form the focus volume. In contrast, 3DED modules apply 3D convolutions directly to the 3D input image stack, resulting in a 3D focus volume that captures both spatial and focus-level information simultaneously. In \citep{yang2022deep}, first multi-scale focus volumes are obtained through the 2D CNN and then the first-order derivatives are applied to get the differential focus volumes. In \citep{won2022learning}, sharp region detection and downsampling modules are suggested for better feature extractions. More recently, \citep{kang2024focdepthformer} introduced FocDepthFormer, which integrates a transformer encoder with an LSTM module and a CNN decoder to enhance global feature reasoning while preserving temporal focus ordering across arbitrary stack lengths. This model leverages pretraining on large-scale monocular depth datasets to incorporate monocular depth cues, thereby improving generalization. Following this, \citep{dogan2025swin} proposed a Swin Transformer-based SFF framework that employs hierarchical windowed self-attention to capture both local and global contextual information across the focal stack. The model integrates multi-scale residual Swin Transformer blocks with a spatial frequency-based focus function, enabling robust focus detection even in low-texture and noisy regions. Unlike conventional CNN-based or hybrid LSTM-CNN architectures, transformer-based SFF models effectively capture long-range dependencies and spatial coherence across focal planes without relying on fixed receptive fields.

Instead of a simple $\arg\max$ approach used in traditional methods, most of the deep learning based SFF techniques map the features to depth through learnable layers. Let  $\mathbf{F} \in  \mathbb{R}^{C \times H \times W}$ be the deep features with $C$ channels, then a simple weight vector ($1 \times 1$ kernel) can be used to map the features to extract a depth map as :
\begin{equation}
\hat{D}(p) \;=\; \sum_{C} \mathbf{\omega} \, . \,\mathbf{F}(p)\,
\end{equation}
where $\mathbf{\omega}$ is the weight vector and its weights are learned during the model training.  In \citep{wang2021bridging}, a different depth extraction approach, motivated by traditional methods, computes a deep focus volume $\mathbf{F} \in \mathbb{R}^{S \times 1 \times H \times W}$ of a single channel that represents focus probabilities for each pixel in the focal stack. This volume is passed through a softplus function and then normalized to provide depth probabilities corresponding to different focus distances. The depth is then extracted through the inner product of the feature vector and the focal distances, which can be expressed as:
\begin{equation}
	\hat{D}(p) \;=\; \sum_{S} \mathbf{\rho} \, . \,\mathbf{F^{\prime}}(p)\,
\end{equation}
where $\mathbf{F^{\prime}}(p)$ are the normalized features and $\mathbf{\rho}$ vector contains the focal distances.  
Current deep learning-based methods place more emphasis on deep feature extraction and less on the depth estimation process. While these approaches leverage large-scale CNNs to capture multi-scale focus cues, they frequently introduce excessive computational complexity without a proportional improvement in depth accuracy.

\subsection{Recurrent Networks in Depth Estimation}

Recurrent Neural Networks (RNNs) are a type of deep learning architecture designed to process sequential data by maintaining an internal memory state. Unlike traditional neural networks, RNNs have connections that form directed cycles, allowing information to persist from one step to the next. They work by taking both the current input and the previous hidden state to generate a new hidden state and output, creating a form of memory that captures information about what has been processed so far in the sequence. Early work such as DepthNet \citep{cs2018depthnet} utilized a convolutional LSTM structure to ensure spatio-temporal coherence when estimating monocular depth from video sequences, and subsequently \citep{yao2019recurrent} demonstrated the use of multiple GRUs for refining depth outputs within multi-view stereo pipelines. The highly regarded Recurrent All-Field Transform (RAFT) techniques \citep{teed2020raft, lipson2021raft} further conceptualized depth refinement as an optimization process, employing multi-scale GRUs to iteratively produce high-fidelity depth fields. In the domain of unsupervised learning, \citep{hui2022rm} developed recurrent modulation units specifically for the iterative and adaptive merging of encoder–decoder features in monocular depth estimation, while more recently Long-short Range Recurrent Updating (LRRU) networks \citep{wang2023lrru} have offered a computationally efficient recurrent approach to progressively enhance an initial depth map.

\begin{figure*}[htb]
        \centering
        \includegraphics[width=\textwidth,keepaspectratio]{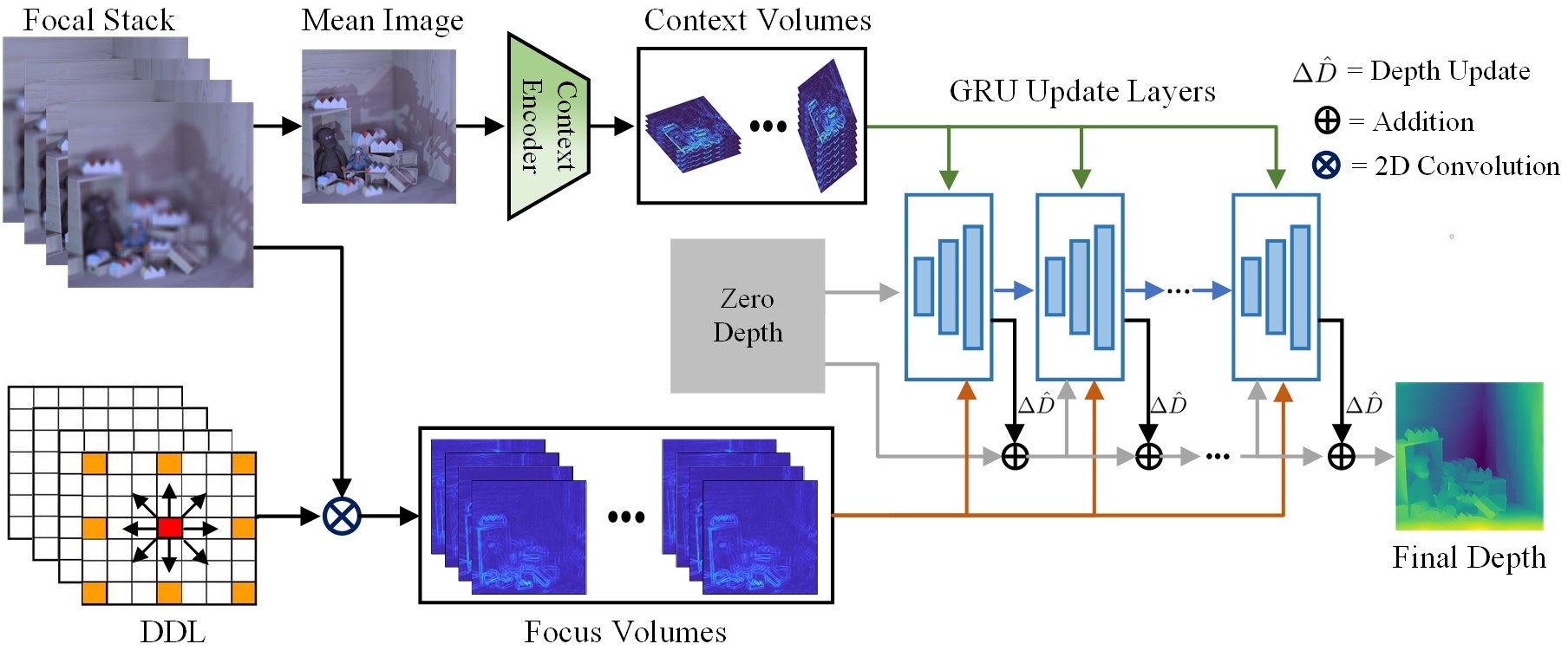}
        \caption{Overview of the proposed pipeline. The method begins by generating multi-scale focus volumes through convolution of the focal stack with Directional Dilated Laplacians (DDL). In parallel, a mean image is extracted from the focal stack and passed through a context encoder to create context volumes. These focus volumes, context volumes, and an initially zero-valued depth map are then provided as inputs to a GRU-based depth extractor, which produces the final depth map.}
	\label{Method}
\end{figure*}

\section{Proposed Method}\label{sec3}
Our proposed method pipeline is shown in \Cref{Method}. The proposed method takes a focal stack as input and outputs a depth map. Initially, multi-scale focus volumes are generated by convolving the focal stack with multiple Directional Dilated Laplacian (DDL) kernels. Simultaneously, a mean image is extracted from the focal stack and passed through a context encoder to produce context volumes. These focus volumes, context volumes, and an initially zero-valued depth map are then fed into a GRU-based depth extractor, which outputs the final depth map.

\subsection{Dilated Laplacian Convolution}\label{sec31}
In dilated convolution, the kernel is \textit{dilated} by inserting spaces (zeros) between its elements, allowing it to cover a larger area. For a 1D input signal $s(i)$, the dilated convolution is defined as:  

\begin{align}
	s'(i) &= \sum_{k=0}^{K-1} w(k) \cdot s(i - r \cdot k)\\
    s'	&=   w \circledast s
\end{align}
where $\circledast$ denotes the convolution between signal $s$  and the kernel $w$ of size $K$, $r$ is the dilation rate (defines the spacing between kernel elements) and $s'$ is the output signal.

The Laplacian kernel is a second-order differential operator that measures the variations from the signal by taking the differences from its neighboring values. In discrete form, the standard 1-D Laplacian and 2D Laplacian for 1D input $s(i)$ is defined as;

\begin{equation}
\scalebox{0.9}{$
\begin{aligned}
\nabla^2 s(i) &= \tfrac{\partial^2s(i)}{\partial i^2}= s(i+1) - 2s(i) + s(i-1).
\end{aligned}
$}
\end{equation}

In conventional SFF methods, a 2D Laplacian kernel is applied and the energy of Laplacian is taken a focus measure. For a 2D input $s_2(i,j)$ it can be expressed as;

\begin{equation}
\scalebox{0.85}{$
\begin{aligned}
\nabla^2 s_2(i,j)
    &= \tfrac{\partial^2 s_2(i,j)}{\partial i^2}
     + \tfrac{\partial^2 s_2(i,j)}{\partial j^2} \\
    &= s_2(i+1,j) + s_2(i-1,j) \\
    &\quad + s_2(i,j+1) + s_2(i,j-1) - 4s_2(i,j)
\end{aligned}
$}
\end{equation}

\begin{figure*}[h]
    \centering
    \includegraphics[width=\textwidth]{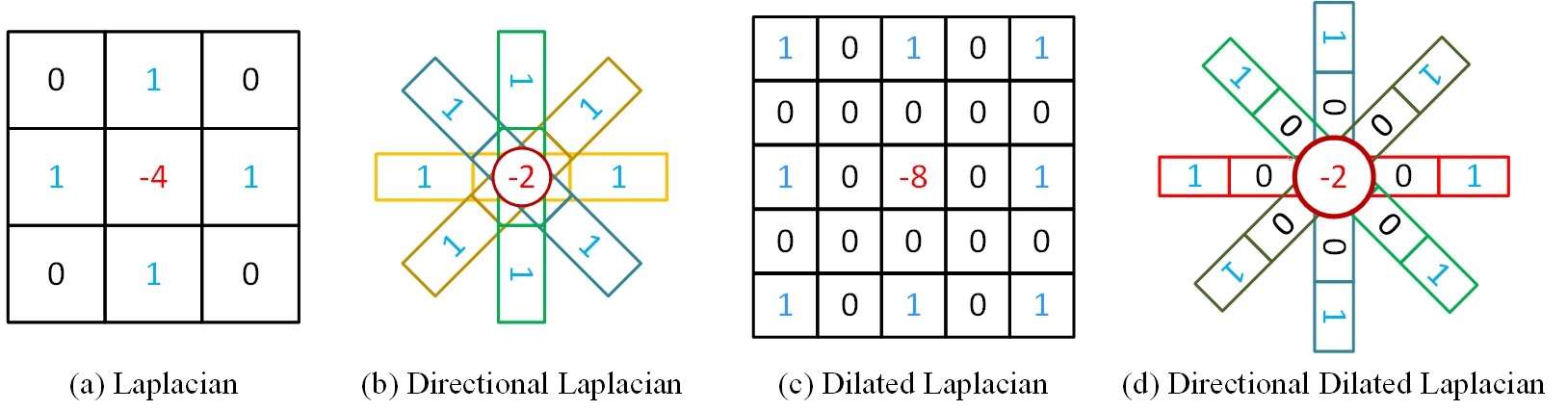}
    \caption{Various kernels: (a) A standard $3 \times 3$ Laplacian, (b) Directional Laplacian, (c) Dilated Laplacian, and (d) Directional Dilated Laplacian (DDL).}
    
    \label{DDL_kernels}
\end{figure*}

There are two major issues when a standard Laplacian kernel is applied for focus computation. The first issue was identified by \citep{nayar1994shape} i.e. in the formulation $|\partial^2(.)/\partial i^2 + \partial^2(.)/\partial j^2|$, partial responses in $i$ and $j$ directions may cancel each other and consequently reduce the focus measure; it was resolved by collecting the energy of each partial derivative, $|\partial^2(.)/\partial i^2| + |\partial^2(.)/\partial j^2|$. Secondly, while Laplacian captures high frequency components  effectively, however, in noisy data, it may amplify the noise and produce inaccurate focus estimates. To tackle this problem, a Ring Difference Filter (RDF) is suggested in \citep{jeon2019ring} by enlarging the perceptive fields by introducing the ring of zeros in the kernel, which is also a form of dilation. However, RDF is also suffering from response cancellation problem. To overcome these limitations, we suggest the Directional Dilated Laplacian (DDL) kernels for focus computation.

  The dilated 1-D Laplacian kernels with different rate $r$ are defined as;
\begin{equation}
	\nabla^2_r s(i) = s(i+r) + s(i-r) - 2s(i).
\end{equation}
Increasing the dilation rate  $r$ allows the operator to detect long-range features. For $r=1$, it is the standard Laplacian kernel, whereas for $r>1$ the dilated Laplacian kernels captures wider-scale variations by skipping intermediate points. In discrete form, the dilated Laplacian kernels can be expressed as;
\begin{align}
\mathcal{H}_{r=1} & = [1, -2, 1] \label{eq:h1} \\
\mathcal{H}_{r=2} & = [1, 0, -2, 0, 1] \label{eq:h2} \\
\mathcal{H}_{r=3} & = [1, 0, 0, -2, 0, 0, 1] \label{eq:h3} \\
\mathcal{H}_{r=4} & = [1, 0, 0, 0, -2, 0, 0, 0, 1] \label{eq:h4}
\end{align}

Different forms of kernels including Laplacian, directional Laplacian, dilated Laplacian and the dilated directional Laplacian are shown in \Cref{DDL_kernels}.

\subsection{Multi-scale Focus Volume Stack }\label{sec32}

Multi-scale focus volumes are obtained by doing convolutions of input focal stack $\mathbf{I}^{(c)}_s(p)$ with a set of directional dilated Laplacian  kernels $ \{\mathcal{H}^{\theta^{\circ}}_r \} $ with  dilation rate $r\in \{1,2,...,R\}$ and the directions $\theta^{\circ} \in \{0^{\circ}, 45^{\circ},90^{\circ}, 135^{\circ}\}$. Focus volume $\mathbf{G}^{(r)}_s(p)$ at dilation rate $r$ is computed by 2D convolution of input images $\mathbf{I}^{(c)}_s(p)$ with 2D kernels for all directions and color channels, then collecting their responses as;
\begin{align}
\mathbf{G}^{(r)}_s(p) \;=\frac{1}{N}\; \sum_{\theta^{\circ}}\sum_{c} | \mathcal{H}^{\theta^{\circ}}_r \,\circledast\, \mathbf{I}^{(c)}_s(p) |^2,
\label{eq:ddl_volumes}
\end{align}
where $\circledast$ denotes the 2D convolution operation. $N=\theta^{\circ} \times c$ is the normalization factor which is product of directions and channels. In other words, an average response from the kernels for all direction for each channel is taken as a focus volume. Thus, repeating this process for all dilation rates, a set of $R$ multi-scale focus volumes $\{\mathbf{G}^{(1)}_s(p),..., \mathbf{G}^{(R)}_s(p)\}$ is obtained.  This set of focus volumes is provided to the GRU-based depth extractor. Note that each of the focus volumes $\mathbf{G}^{(r)}_s(p) \in \mathbb{R}^{S \times 1\times H \times W}$  possess the same spatial dimensions as of input stack  $\mathbf{I}_s(p)$. In practice we are computing four of these focus volumes, hence $R=4$ for our experiments.

\begin{figure*}[htb]
        \centering
        \includegraphics[width=\textwidth,keepaspectratio]{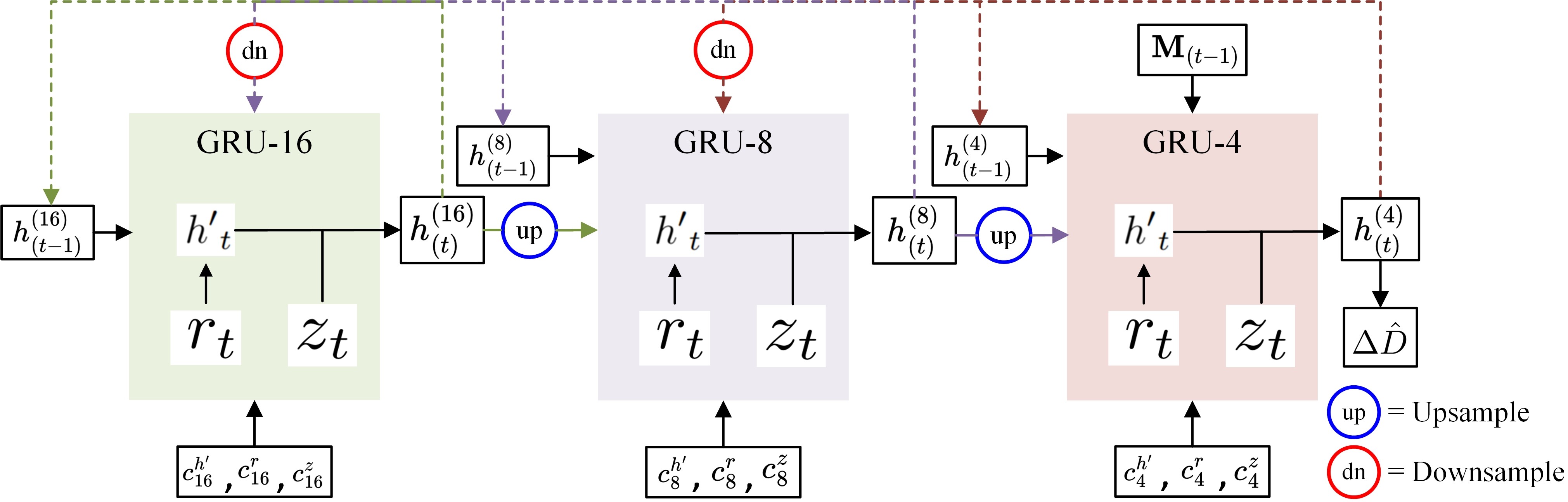}
        \caption{An overview of our proposed GRU-based depth extraction module working across different scales at iteration $t$. GRU-16 is the coarsest scale, GRU-8 is the intermediate and GRU-4 is the finest scale that is eventually responsible for evaluating depth update. The information flows between these adjacent layers as shown.}

	\label{GRU}
\end{figure*}

\subsection{Context Encoder} \label{sec33}

Context Encoder module takes the reference image and provides a set of multi-scale global context features.  Architecturally, our context encoder employs a multi-stage convolutional backbone with residual connections to progressively extract hierarchical features. The network begins with initial convolution and normalization layers, followed by a series of residual blocks that incrementally increase the number of channels while controlling spatial resolution through downsampling operations. Among the various choices for reference image, we choose the mean image which is a simple yet effective choice to guide the context encoder for capturing scene priors. As observed in earlier work \citep{ali2021guided}, mean image provides rich contextual information about the scene and plays an important role in optimizing the energy functional for enhancing the depth map. The mean image $\bar{\mathbf{I}}^{(c)}(p)$ from the focal stack is computed as:
\begin{equation}
	\bar{\mathbf{I}}^{(c)}(p)=\frac{1}{S}\sum_s\mathbf{I}^{(c)}_{s}(p).
\end{equation}

The context encoder produces three gate-specific multi-scale bias volumes per ConvGRU scale, $c_q^{z}$, $c_q^{r}$ and $c_q^{h'}$, for $q\in\{4,8,16\}$, where $q$ denotes the downsampling factor (scale) at which the ConvGRU operates. Each bias volume has shape $\mathbb{R}^{128\times (H/q)\times (W/q)}$. These bias volumes initialize the hidden state at the corresponding ConvGRU scale and also provide a global prior at each iteration by being injected additively into the update ($z$), reset ($r$) and candidate ($h'$) convolution outputs of the GRU. The choice of $q$ is configurable and may be changed to suit different resolution requirements.

\subsection{Depth Extraction Module}\label{sec34}

The next stage of our method is an iterative, multi-scale depth extraction module inspired by the proven success of recurrent refinement strategies in optical flow \citep{teed2020raft} and stereo matching \citep{lipson2021raft}. Instead of generating the depth map in a single pass, the network progressively refines an initial estimate by learning error-correcting update rules over multiple iterations. Although other regression structures could be employed, we adopt the GRU-based formulation as an effective choice motivated by its demonstrated success in optical flow and stereo matching.

\textbf{Iterative Refinement:} Our GRU-based structure operates at multiple scales, as illustrated in \Cref{GRU}. We denote the GRU layer at the coarsest level as GRU-16, the intermediate level as GRU-8, and the finest level as GRU-4. Moreover, the finest level (GRU-4) is responsible for updating the depth map. The initial depth map is set to zero: $\hat{D}_{0}=0$. We construct a focus aggregation map $\mathbf{U} \in \mathbb{R}^{(R \cdot S) \times 1 \times H \times W}$ by stacking focal slices extracted from all focus volumes $\mathbf{G}^{(r)}_s(p)$. This map collects sharpness evidence from all $S$ images across all $R$ dilation rates. At each iteration $t$, fused focus–depth features are prepared exclusively for the finest GRU level by concatenating the previous (or initialized) depth $\hat{D}_{t-1}$ with $\mathbf{U}$, followed by a small stack of convolutions to produce $\mathbf{M}_{(t-1)}$. The GRU update at iteration $t$ can then be written compactly as:

\begin{equation}
    h_{(t)} = \mathrm{GRU}\bigl(h_{(t-1)},\, {b}_{(t)},\,  c^{z},\, c^{r},\, c^{h'}\bigr),
\end{equation}

where $b_{(t)}$ is an auxiliary input specific to each GRU scale. For clarity, we omit scale indices in the equations. The update relies on a convolution-based gating mechanism, adapted to our SFF design, and proceeds as:

\begin{align}
	z_t &= \sigma\bigl(\mathrm{Conv}_{3\times3}([h_{(t-1)},\, b_{(t)}],\, W_z) + c^{z} \bigr), \label{eq:convgru_1}\\
	r_t &= \sigma\bigl(\mathrm{Conv}_{3\times3}([h_{(t-1)},\, b_{(t)}],\, W_z) + c^{r} \bigr), \label{eq:convgru_2}
\end{align}
\vspace{-30pt}
\begin{align}
    h'_t &= \scalebox{0.83}{%
    $\tanh\!\Bigl(\mathrm{Conv}_{3\times3}\bigl([r_t \odot h_{(t-1)},\, b_{(t)}],\, W_h\bigr) + c^{h'}\Bigr),$%
    }\label{eq:convgru_3}
\end{align}
\vspace{-30pt}
\begin{align}
    h_t &= (1 - z_t) \odot h_{(t-1)} \;+\; z_t \odot {h'}_t,
	\label{eq:convgru_4}
\end{align}

where $z_t$ and $r_t$ are the update and reset gates, respectively; $\odot$ denotes element-wise multiplication; $\mathrm{Conv}_{3\times3}$ indicates a $3 \times 3$ convolution; $W_z$, $W_r$, and $W_h$ are learnable convolutional weights; $\sigma$ is the sigmoid function; and $\tanh$ is the hyperbolic tangent.

The auxiliary input $b_{(t)}$ incorporates information from neighbouring GRU scales, while the finest scale (GRU-4) also uses fused focus–depth features $\mathbf{M}_{(t-1)}$. Our multi-scale GRU architecture is organized hierarchically, as shown in \Cref{GRU}. The coarsest GRU receives features at its resolution along with pooled information from the intermediate scale. The intermediate GRU is fed its own features, a pooled representation from the finer scale, and an upsampled hidden state from the coarsest GRU. Finally, the finest GRU processes its own features together with an upsampled hidden state from the intermediate layer, as well as the fused features $\mathbf{M}_{(t-1)}$. Finally, a lightweight depth prediction layer comprising two convolutions and a ReLU uses the finest-scale hidden state to predict a depth update $\Delta \hat{D}$ that is added to the previous depth as:
\begin{equation}
	\hat{D}_t = \hat{D}_{t-1} + \Delta \hat{D}
\end{equation}
This update is applied iteratively over  $T$ steps, enabling the network to progressively correct depth estimation errors and incorporate additional context at each iteration. Our final depth after $T$ iterations is then $\hat{D}_T$.

\textbf{Learned Convex Upsampling:} Since the GRU refinement operates on downsampled representations, a learned upsampling module is applied to recover the full $ H \times W $ resolution. Specifically, the updated hidden state at the finest level is used to predict a set of 9 weights per pixel for a $ 3 \times 3 $ neighborhood in the high-resolution grid.  Each set of weights defines a convex combination of the 9 nearest neighbors, effectively reconstructing a refined high-resolution depth map. This learned upsampling preserves fine details more effectively than traditional interpolation methods.

\subsection{Loss} \label{sec35}
Each intermediate depth estimate $\hat{D}_{t}$ is supervised during training with the GT depth map ${D}'$ using a mean squared error with progressively increasing weights as :
\begin{equation}
	\mathcal{L} = \sum_{t=1}^{T} \alpha^{T-t} \left( {D}' -\hat{D}_{t}  \right)^2,
\end{equation}
where we set $\alpha = 0.9$ for our experiments.

\section{Results and Discussions}\label{sec4}

\subsection{Experimental Setup}\label{sec41}

\subsubsection{Training}
Our model comprises a total of around 10.04M trainable parameters, of which 4.31M are allocated to the context encoder and the remaining 5.73M belong to the GRU-based depth extraction module. The initial weights of both modules were randomly initialized. The entire implementation is performed in PyTorch~\citep{paszke2019pytorch}. The model is optimized using the Adam optimizer~\citep{kingma2014adam} with parameters $\beta_1 = 0.9$ and $\beta_2 = 0.999$. Training is conducted on patches of size $256 \times 256$, with a starting learning rate of $1\times10^{-4}$ that is reduced after a fixed number of epochs, depending on the training dataset. Our architecture employs three GRU layers operating at resolutions of $\frac{1}{16}$, $\frac{1}{8}$, and $\frac{1}{4}$ of the full input resolution. The refinement process initiates at the coarsest level and propagates to the finest level, with the finest level responsible for predicting $\Delta \hat{D}_t$. Additionally, we perform 32 iterative updates during the GRU-based recurrent refinement process. For simplicity, we have converted our focal stacks to grayscale before they are fed to the model. Notably, our model is designed to be generalizable, allowing training and testing on focal stacks with any number of images, independent of the dataset used for training.

\subsubsection{Datasets}  We conducted our experiments on a total of six datasets. Of these, five diverse datasets collectively including over 1,600 focal stacks and more than 19,500 images with varying focal points were used to analyze and compare model performance. The sixth dataset, LFSD, was specifically used for analyzing dilated Laplacians.

\begin{itemize}
    \item \textit{FlyingThings3D (FT): } \citep{mayer2016large}: A large-scale synthetic dataset comprising 1000 focal stacks for training and 100 for testing. Each stack contains 15 images captured at different focal settings with focus distances evenly distributed from 10 to 100 units.
    
    \item \textit{FocusOnDefocus (FoD): } \citep{maximov2020focus}: A synthetic dataset consisting of 500 focal stacks, each with 5 images at a resolution of $256 \times 256$. Out of these, 400 stacks are allocated for training and 100 for testing. The focal distances for the images are $\{0.1, 0.15, 0.3, 0.7, 1.5\}$ units.
    
    \item \textit{Middlebury: } \citep{scharstein2014high}: A real-world dataset originally intended for stereo tasks and repurposed here for depth-from-focus. It contains 15 focal stacks, each with 15 images. Due to its limited size, it is used exclusively as a test set, with images resized to $512 \times 512$. The focus distances are evenly distributed between 10 and 60 units.
    
    \item \textit{Mobile: } \citep{suwajanakorn2015depth}: A real-world dataset captured using a Samsung Galaxy S3 mobile phone. It consists of 13 aligned focal stacks, each containing between 12 and 33 images captured at varying focal settings.
    
    \item \textit{HCI: } \citep{honauer2017dataset}: A synthetic dataset derived from 4D light fields, containing 24 focal stacks with 10 images each. Out of these, 20 stacks are used for training and 4 for testing. Given its limited number of focal stacks, the HCI dataset is primarily utilized for ablation studies. For experiments focused on traditional depth extraction, 30 focus varying images were generated from 14 focal stacks within the original dataset using the tools from \cite{dansereau2015light}.

    \item \textit{LFSD: } \citep{li2014saliency} This is a real-world dataset captured using a Lytro light field camera. It contains 100 scenes, split into 60 indoor and 40 outdoor scenes, with 3 to 12 focus-varying images per scene. This dataset was solely used to test the robustness of the proposed DDL. The depth maps provided by the authors are used as pseudo-ground truth.
    
\end{itemize}

\subsection{Ablation Study}\label{sec42}

\begin{figure}[htbp]
    \centering
    \includegraphics[width=\columnwidth,keepaspectratio]{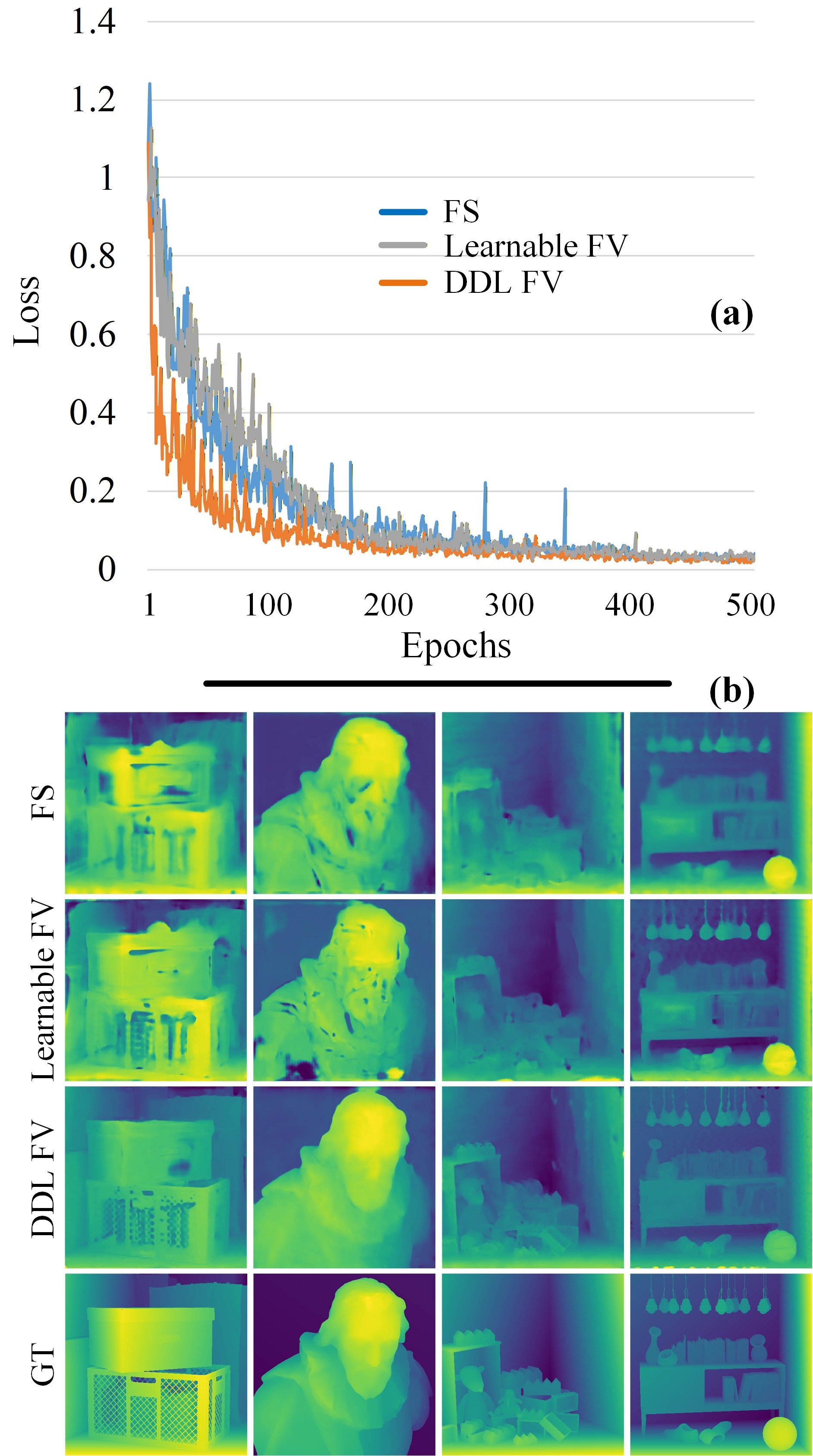}
    \caption{(a) Training loss comparison for three inputs (FS / Learnable FV / DDL FV) showing faster and smoother convergence with DDL FV. (b) Outputs for the three inputs: DDL input yields the sharpest and cleanest depth predictions.}
    \label{losscurve}
\end{figure}

\subsubsection{Input Representations}
One of the key findings of our work is that providing explicit, handcrafted focus features to a deep model significantly improves training efficiency and final depth quality compared to using the raw focal stack (FS). To validate this, we trained the AiFDNet \citep{wang2021bridging} method on the HCI dataset under identical settings using three different inputs: (i) the original focal stack (FS), (ii) a focus volume constructed by a stack of learnable dilated convolutional kernels with receptive fields and dilation rates matched to DDL (`learnable FV'), and (iii) our proposed DDL focus volume, computed from a $3 \times 3$ kernel in four directions for each color channel separately, provided as a three-channel input. As shown in \Cref{losscurve}(a), the model trained with the handcrafted DDL FV converges faster and more smoothly than both the FS and learnable-FV variants. Notably, FS and learnable FV exhibit frequent fluctuations and sharp loss variations even after 300 epochs, whereas DDL maintains a much steadier convergence with fewer such peaks, indicating more stable training behavior.

Qualitative results reinforce the loss-curve observations. \Cref{losscurve}(b) shows output maps for the three input types: the DDL-based input yields the cleanest output predictions with the sharpest object boundaries, clearer separation between near and far objects (with distant objects appearing distinctly farther), and markedly reduced high-frequency noise. The learnable-FV, despite having the capacity to adapt during training, produces outputs that are generally noisier and less accurate than those produced from DDL; this suggests that without an explicit focus cue, the network struggles to reliably learn the high-frequency focus responses needed for accurate results. Moreover, since most SFF datasets lack ground-truth FV, the model’s effectiveness depends heavily on the quality of the FV. Explicitly encoding high-frequency directional focus cues through DDL proves more effective than relying on the network to learn them implicitly.

\begin{figure*}[htbp] 
    \centering
    \includegraphics[width=\textwidth,keepaspectratio]{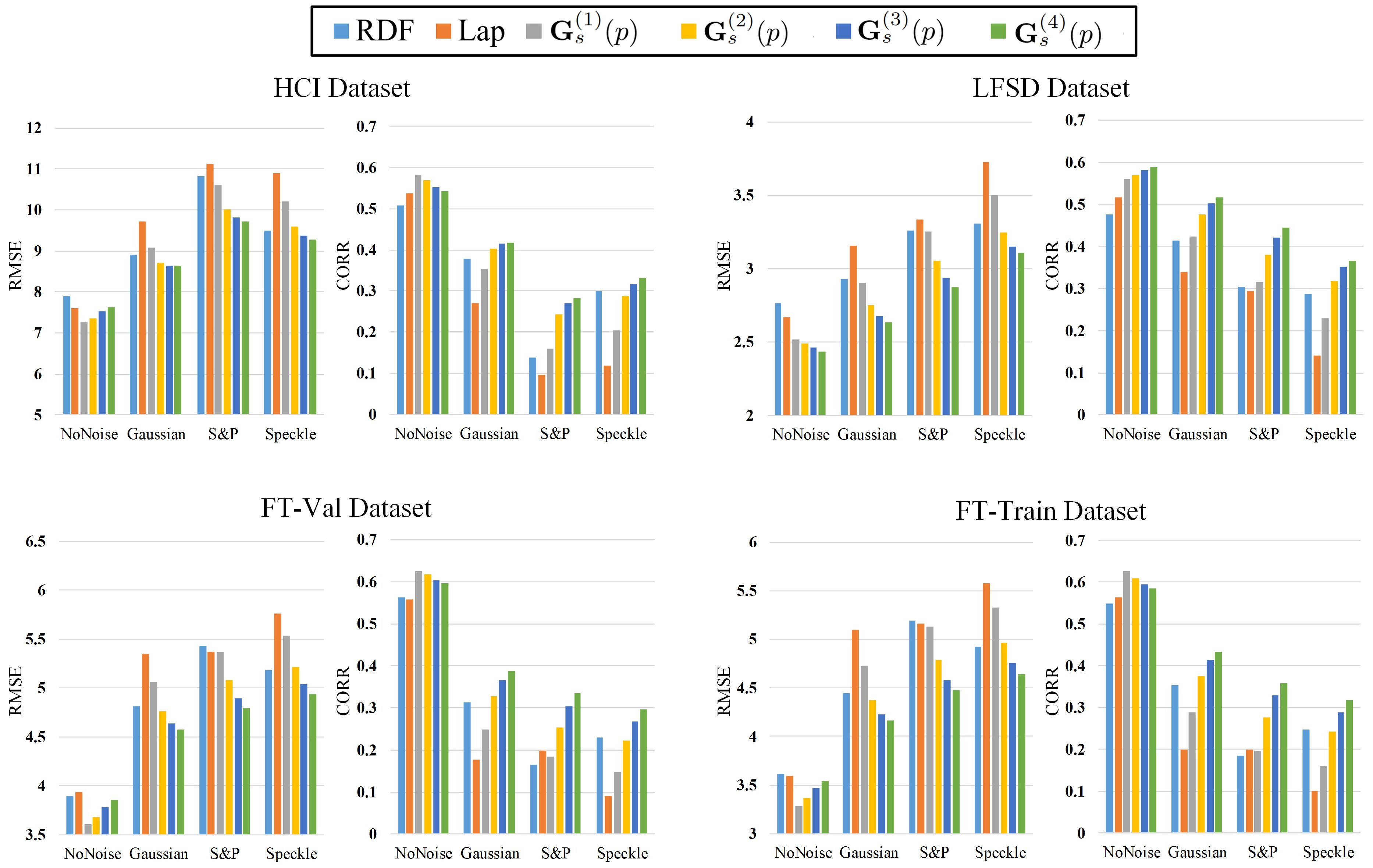} 
    \caption{Noise robustness comparison using RMSE (lower better) and CORR (higher better). Performance of baselines RDF and Laplacian (Lap) vs. progressively added DDL variants ($\mathbf{G}^{(1)}_s(p)$ to $\mathbf{G}^{(4)}_s(p)$, under no noise and also with Gaussian, S\&P, and Speckle noises. Results across evaluated datasets show improved resilience with DDL, especially more directional responses of a bigger dilated kernel (increasing $r$) are added.}
	\label{Dil_Lap_Quant} 
\end{figure*}

\subsubsection{Robustness}\label{secA2}

To evaluate the robustness of our proposed DDL, against noise and to validate our choice of using multi-scale DDL volumes, we conducted a comparative experiment. We assessed depth estimation performance under four different conditions: noise-free, Gaussian noise (mean 0, variance 0.0001), Salt and Pepper (S\&P) noise (density 0.005, i.e. 0.5\%), and Speckle noise (variance 0.005), applied to the input images before processing. As baselines, we used the well-established Ring Difference Filter (RDF) \citep{jeon2019ring}  and the standard 3x3 Laplacian operator (Lap). These were compared against four variants of our DDL computed FVs: $\mathbf{G}^{(1)}_s(p)$, $\mathbf{G}^{(2)}_s(p)$, $\mathbf{G}^{(3)}_s(p)$, and $\mathbf{G}^{(4)}_s(p)$. All comparative methods, including baselines, were used to compute the FV and then extract depth using the traditional SFF method. It is to be noted that the DDL variants used in this experiment were designed progressively, such that each subsequent variant $\mathbf{G}^{(r+1)}_s(p)$ integrates additional directional responses derived from an incrementally larger dilation rate $r+1$ into the FV computation, building upon the responses included in the previous variant $\mathbf{G}^{(r)}_s(p)$. This progressive addition was designed because traditional SFF methods operate on a single focus volume to extract depth; hence, to analyze the effect of incorporating multiple scales, we incrementally added responses from previous dilation levels to observe the impact of multi-scale integration. The evaluation was performed across four datasets, \textit{HCI}, \textit{LFSD}, and \textit{FT} (treating the training and test splits of the \textit{FT} dataset as distinct), encompassing a total of over 1,200 focal stacks and more than 17,000 images.

Quantitative results, measured by Root Mean Square Error (RMSE) and Correlation Coefficient (CORR), are presented in \Cref{Dil_Lap_Quant}. The findings demonstrate two key points. Firstly, incorporating directional information significantly enhances robustness over the standard Laplacian; even the simplest DDL variant, $\mathbf{G}^{(1)}_s(p)$, often outperformed the baseline Lap operator across tested noise conditions and also often outperformed RDF, for example, when observing CORR on the \textit{FT-Val} dataset under Gaussian noise conditions. Secondly, there is a clear trend of improved depth estimation accuracy as more directional responses from larger dilation rates are added. This is evidenced by comparing successive DDL variants: $\mathbf{G}^{(r+1)}_s(p)$ consistently yields better quantitative metrics (lower RMSE, higher CORR) than $\mathbf{G}^{(r)}_s(p)$, indicating that leveraging information from larger DDL kernels improves noise resilience and depth quality. However, the noise-free case serves as a special condition where simply enlarging the receptive field with higher dilations degrades performance in synthetic datasets such as \textit{HCI} and \textit{FT}, as seen when comparing the performance of $\mathbf{G}^{(1)}_s(p)$ with $\mathbf{G}^{(2)}_s(p)$, which oversmooths fine details. In contrast, this effect is not observed in the real-world \textit{LFSD} dataset, where incorporating larger dilation rates actually improves performance even under the noise-free condition. We believe this difference arises because real-world captures inherently contain slight sensor or illumination noise, where broader receptive fields help produce more stable and reliable responses. To balance both synthetic and real-world conditions, we selected a dilation rate of $r = 4$, which provides strong robustness in practical scenarios while maintaining high accuracy in clean, noise-free cases.

\begin{figure*}[h]
    \centering
    \includegraphics[width=\textwidth,keepaspectratio]{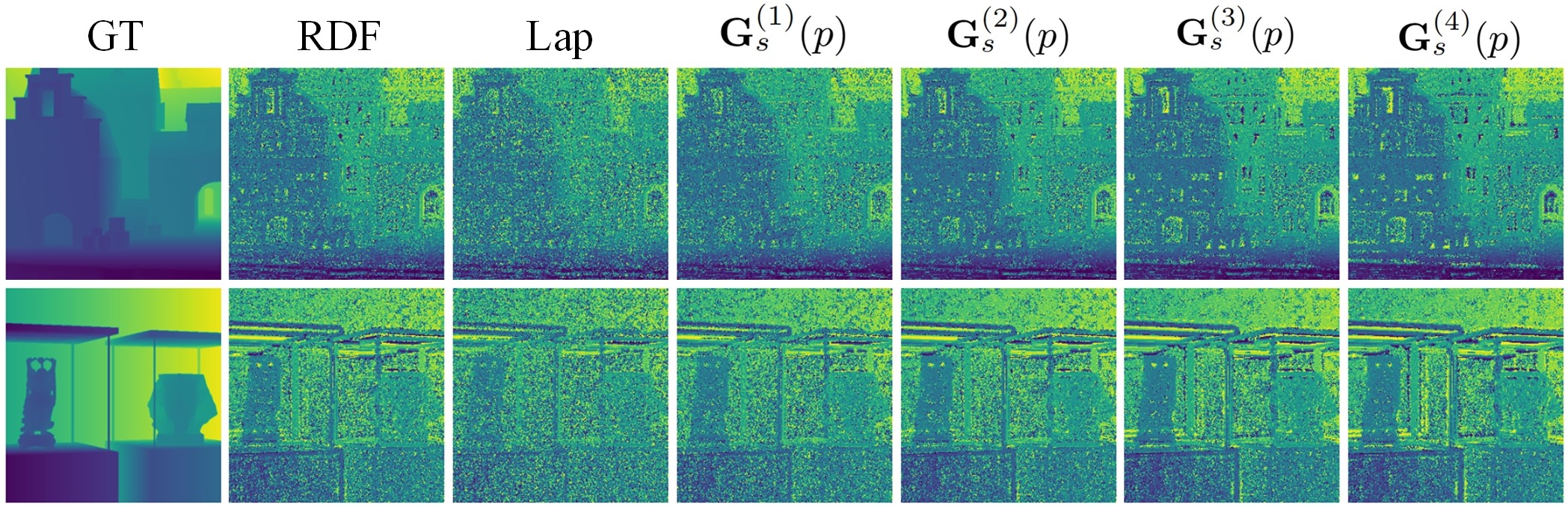}
    \caption{Qualitative comparison of depth maps generated using different focus volumes on the \textit{HCI} dataset.}
    \label{Dil_Lap_Depths}
\end{figure*}

A qualitative assessment was performed to compare the resulting depth maps. For this, an evaluation was conducted on images from the \textit{HCI} dataset corrupted by Gaussian noise (mean 0, variance 0.0001). The performance of a standard $3 \times 3$ Laplacian operator (Lap), together with RDF \citep{jeon2019ring}, served as our baseline; these were compared against four progressively added variants of our DDL-computed focus volumes: $\mathbf{G}^{(1)}_s(p)$, $\mathbf{G}^{(2)}_s(p)$, $\mathbf{G}^{(3)}_s(p)$, and $\mathbf{G}^{(4)}_s(p)$. The depth maps were computed using the traditional approach outlined in \Cref{depth_trad} (see quantitative RDF results in \Cref{Dil_Lap_Quant}).

The resulting depth maps are shown in \Cref{Dil_Lap_Depths}. As can be observed, the Lap approach performs the worst. RDF qualitatively reduces some high-frequency noise relative to Lap and produces more coherent depth regions, but it still exhibits more residual noise and less well-defined object boundaries than the stronger DDL variants. Robustness to noise and the quality of the depth maps improve as we progress from $\mathbf{G}^{(1)}_s(p)$ to $\mathbf{G}^{(4)}_s(p)$.

\subsubsection{Focus Measure Curve Analysis}

A focus measure (FM) curve illustrates the focus of a given pixel across each frame within a focus volume (FV) by plotting the focus value against the image index. The peak of the FM curve indicates the image index corresponding to the maximum sharpness for that pixel. Ideally, an FM curve should exhibit a single, distinct peak, demonstrating the robustness of the focus measure operator in confidently identifying focus values for a pixel while mitigating the impact of noise.

\begin{figure*}[hbtp]
        \centering
        \includegraphics[width=\linewidth,keepaspectratio]{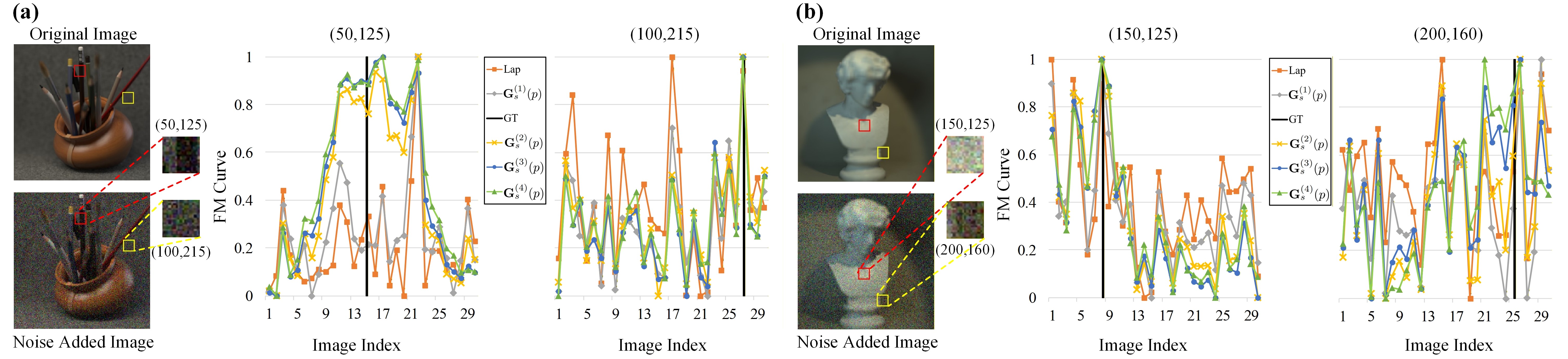}
        \caption{(a)\,FM curve analysis on the \textit{Pens} focal stack and
        (b)\,on the \textit{Antinous} from the \textit{HCI} dataset. 
        Note that noise has been enhanced for visualization purposes.}
	\label{FM_Curve_Combined}
\end{figure*}

Specifically, we assessed the performance of our approach on images from the \textit{HCI} dataset corrupted by Gaussian noise (mean 0, variance 0.0001). A standard 3x3 Laplacian operator (Lap) served as our baseline. We compared this baseline against four progressively added variants of our DDL-computed FVs: $\mathbf{G}^{(1)}_s(p)$, $\mathbf{G}^{(2)}_s(p)$, $\mathbf{G}^{(3)}_s(p)$, and $\mathbf{G}^{(4)}_s(p)$ as described in \Cref{secA2}. This analysis was conducted on a total of four randomly selected pixels, two from the \textit{Pens} focal stack and two from the \textit{Antinous} focal stack, both from the \textit{HCI} dataset.

The FM curves for the \textit{Pens} scene are presented in \Cref{FM_Curve_Combined}(a) for two pixels: (50,125) and (100,215). In these curves, the y-axis represents the focus value, while the x-axis indicates the image index at which the focus value is plotted. The ground truth (GT) value for each pixel's depth is also indicated. As shown in the figure, incorporating more dilated responses into the focus estimation results in the FM curve's peak aligning more closely with the GT value. Notably, all four DDL variants, $\mathbf{G}^{(1)}_s(p)$, $\mathbf{G}^{(2)}_s(p)$, $\mathbf{G}^{(3)}_s(p)$, and $\mathbf{G}^{(4)}_s(p)$, were relatively better than the standard Laplacian operator.

A similar trend is observed for two pixels from the \textit{Antinous} scene, as illustrated in \Cref{FM_Curve_Combined}(b). In this case, $\mathbf{G}^{(4)}_s(p)$ performs the best, exhibiting the closest alignment with the GT value.

\subsubsection{Deep Encoders vs DDL}

To validate the effectiveness of DDL as an effective FV, in this experiment, we utilize deep neural encoders to extract FV that are subsequently fed into our proposed GRU-based depth extraction module. Deep encoders produce multi-scale, channel-dense feature volumes, where each successive volume exhibits a reduction in the number of channels along with an increase in spatial resolution. To remain consistent with classical SFF theory, where the FV is defined as a single-channel, per-pixel scalar response, we transform these channel-dense feature volumes into single-channel FVs using a lightweight collapser module. The collapser consists of two convolutional layers: the first projects the channel-dense volume into a 32-channel representation, and the second reduces it to a single-channel volume. Finally, the output is upsampled to match the original resolution. While this single-channel formulation provides clearer per-pixel focus cues and better interpretability in line with SFF, it is not a strict requirement, and our GRU-based depth extractor can in principle also operate on multi-channel feature volumes. We evaluated four neural encoders: VGG-16 \citep{simonyan2014very}, ResNet-18 \citep{he2016deep}, MobileNetV3 \citep{howard2019searching}, and the Atto version of ConvNeXtV2 \citep{woo2023convnext} and compared them with the propsed DDL. For a fair comparison, we used the \textit{FT} dataset for evaluation, where each was trained with a batch size of 2 and 8 GRU iterations. Table~\ref{Neural_Encoder_Quant} presents the total additional parameter counts for each encoder, which include both the parameters of the encoder itself and the collapser when integrated with our GRU-based depth extraction module, along with the quantitative evaluation. While ConvNeXtV2 achieved the best overall performance, as expected from a recent state-of-the-art model, the proposed DDL method, despite having no learnable parameters, produced competitive results and even ranked second-best in some metrics.

\begin{table}[htbp]
  \centering
  \caption{Additional parameter count and quantitative evaluation for different neural encoders for \textit{FT} dataset when used with our proposed GRU-based depth extraction module. 'DDL' is not a neural encoder but represents our proposed traditional FV computation method.}
   \label{Neural_Encoder_Quant}%
   \footnotesize
   \setlength{\tabcolsep}{0.45\tabcolsep}
    \begin{tabular}{cccccc}
    \toprule
    Encoder & Params & MAE   & AbsRel & Acc\_1.25 & Acc\_1.25$^3$ \\
    \midrule
    VGG-16 & 7.7 M & 8.99  & 0.28  & 64.81 & 95.57 \\
    ResNet-18 & 11.2 M & 2.10  & 0.08  & 95.42 & 99.05 \\
    MobileNetV3 & 3.0 M & 2.10  & 0.10  & 95.10 & 98.84 \\
    ConvNeXtV2 & 3.4 M & 2.03  & 0.07  & 96.21 & 99.26 \\
    DDL   & 0     & 2.39  & 0.08  & 95.73 & 99.25 \\
    \bottomrule
    \end{tabular}%
 
\end{table}%

\begin{figure*}[h]
        \centering
        \includegraphics[width=\textwidth,keepaspectratio]{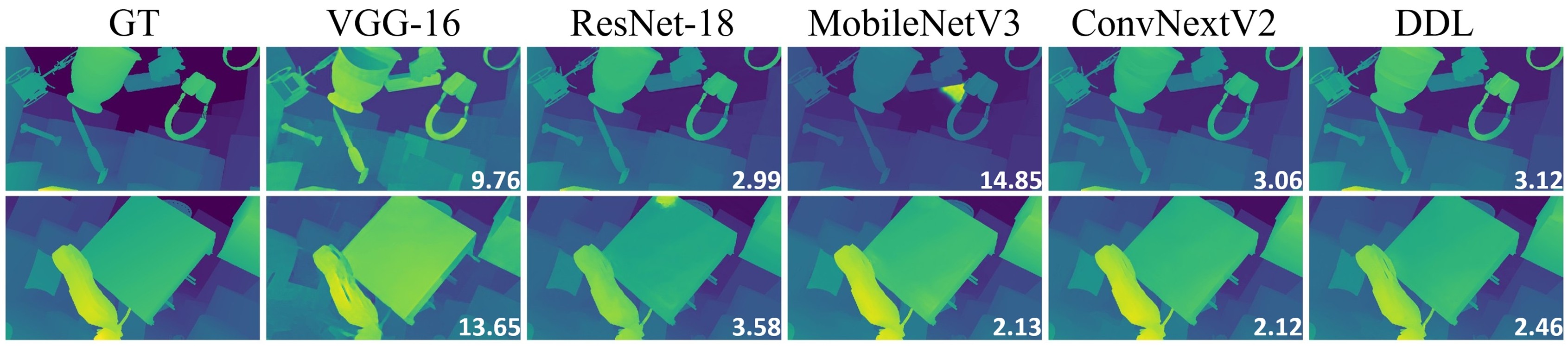}
        \caption{Qualitative comparison on \textit{FT} dataset using different neural encoders when used with our proposed GRU-based depth extraction module. Numbers on each map denote the corresponding RMS error with respect to the GT.}

	\label{Neural_Encoder_qual}
\end{figure*}

Fig.~\ref{Neural_Encoder_qual} shows qualitative comparisons. All methods perform reasonably well, with VGG-16 and MobileNetV3 showing slightly lower quality when observing outputs from the first row, and ResNet-18 also showing slightly lower quality when observing outputs from the second row, while ConvNeXtV2 and DDL produce relatively clearer outputs for both, which is also reflected in the corresponding RMS error values shown on each map. These findings demonstrate that DDL can effectively replace learned FV computation modules without a notable drop in performance.

\subsubsection{GRU-Based Regression vs Alternatives}

To evaluate the effectiveness of the proposed GRU-based depth extraction module, we conducted a controlled experiment where the same DDL-based focus volume (FV) was provided as input to different state-of-the-art regression frameworks, namely AiFDNet \citep{wang2021bridging} and DFV-FV \citep{yang2022deep}. AiFDNet is a U-Net–style 3D encoder–decoder model, while DFV-FV is a hybrid 2D–3D encoder–decoder that first extracts 2D features using a ResNet backbone, then builds and decodes 3D feature volumes through multi-scale fusion and skip connections. Since these networks are designed to take a single three-channel input, we computed DDL responses from a $3 \times 3$ kernel in four directions for each color channel separately and provided them as a three-channel input. All experiments were performed on the \textit{FT} dataset. Quantitative results are summarized in \Cref{Regressor_Ablation_Quant}. The proposed GRU-based regression head achieves the best overall performance, benefiting from its recurrent refinement mechanism that iteratively corrects residual errors across time steps.

\begin{table}[htbp]
  \centering
  \caption{Comparison of GRU and alternative depth regression modules using DDL-based FV as input. Bold denotes the best performance.}
    \setlength{\tabcolsep}{0.5\tabcolsep}
    \begin{tabular}{cccccc}
    \toprule
    Regressor & MAE   & MSE   & AbsRel & Acc\_1.25 & Acc\_1.25$^3$ \\
    \midrule
    AiFDNet & 5.45  & 368.21 & 0.61  & 86.34 & 90.04 \\
    DFV-FV & 5.58  & 370.50 & 0.62  & 86.03 & 89.96 \\
    Ours-GRU & 2.70  & 102.17 & 0.10  & 95.38 & 99.09 \\
    \bottomrule
    \end{tabular}
  \label{Regressor_Ablation_Quant}%
\end{table}

\begin{figure}[H]
        \centering
        \includegraphics[width=\columnwidth,keepaspectratio]{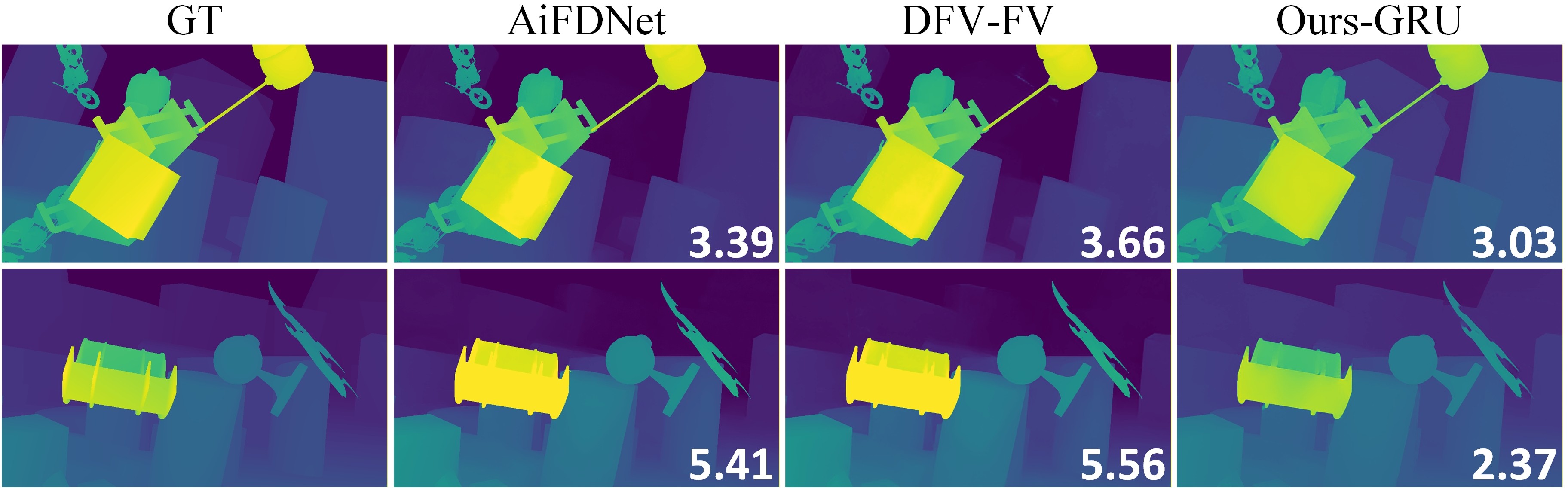}
        \caption{Qualitative comparison of GRU and alternative depth regression modules using DDL-based FV as input. Numbers on each map denote the corresponding RMS error with respect to the GT.}

	\label{Regressor_Ablation_Qual}
\end{figure}

Representative qualitative results are shown in \Cref{Regressor_Ablation_Qual}, where the numbers on each map indicate the corresponding RMS error with respect to the GT. The advantages of the GRU-based regressor are particularly noticeable in the background regions, where its iterative correction helps preserve structural details such as the building-like structures visible in the GT that other regressors often fail to capture.

\subsection{Comparative Analysis}
\subsubsection{Comparative Methods}
We conducted a comprehensive evaluation of our method against a wide range of state-of-the-art approaches across multiple benchmark datasets, including \textit{FT}, \textit{Middlebury}, \textit{FoD}, and \textit{Mobile}. On the \textit{FT} and \textit{Middlebury} datasets, we compared our method with both traditional and deep learning-based techniques. This includes RFVR \citep{ali2021robust}, a conventional regularization-based method, and several learning-based models: AiFDNet \citep{wang2021bridging}, DWild \citep{won2022learning}, and the DFV family of models \citep{yang2022deep}, including DFV-FV (the base version) and DFV-Diff (which introduces a differential focus volume approach). For the \textit{FoD} dataset, we benchmarked against RFVR \citep{ali2021robust}, AiFDNet \citep{wang2021bridging}, and DFV-FV and DFV-Diff \citep{yang2022deep}. On the \textit{Mobile} dataset, we include DDFF \citep{hazirbas2019deep}, DefocusNet \citep{maximov2020focus}, AiFDNet \citep{wang2021bridging}, DFV-FV and DFV-Diff \citep{yang2022deep}, and DDFS \citep{fujimura2024deep} for comparison.

\begin{table*}[htbp]
  \centering
  \caption{Quantitative results on the FT dataset. Bold indicates the best performance.}
  \resizebox{1\textwidth}{!}{%
    \begin{tabular}{ccccccccccccc}
    \toprule
    Method & MAE   & MSE   & RMS   & logRMS & AbsRel & SqRel & Acc\_1.25 & Acc\_1.25$^2$ & Acc\_1.25$^3$ & BadPix & Bump & CORR \\
    \midrule
    RFVR  & 11.89 & 1035.34 & 23.63 & 0.79  & 1.55  & 82.66 & 72.79 & 80.81 & 84.60 & 98.54 & 4.12  & 0.73 \\
    AiFDNet & 6.81  & 642.05 & 13.14 & 0.59  & 0.73  & 11.30 & 85.43 & 87.67 & 88.87 & 90.37 & 4.66  & 0.93 \\
    DFV-FV & 6.33  & 427.58 & 12.09 & 0.57  & 0.90  & 27.04 & 85.09 & 87.60 & 89.52 & 92.43 & 3.99  & 0.94 \\
    DFV-Diff & 5.51  & 376.72 & 10.65 & 0.53  & 0.62  & 7.48  & 86.18 & 88.09 & 89.93 & \textbf{89.82} & \textbf{3.85} & 0.97 \\
    DWild & 5.54  & 392.00 & 10.44 & 0.53  & 0.61  & 7.24  & 86.35 & 88.21 & 89.84 & 90.02 & 4.29  & 0.97 \\
    Ours  & \textbf{2.70} & \textbf{102.17} & \textbf{6.26} & \textbf{0.16} & \textbf{0.10} & \textbf{4.13} & \textbf{95.38} & \textbf{98.38} & \textbf{99.09} & 91.51 & 4.90  & \textbf{0.98} \\
     \bottomrule
    \end{tabular}%
    }
  \label{FT_Quant}%
\end{table*}%

\begin{figure*}[htbp]
        \centering
        \includegraphics[width=\textwidth,keepaspectratio]{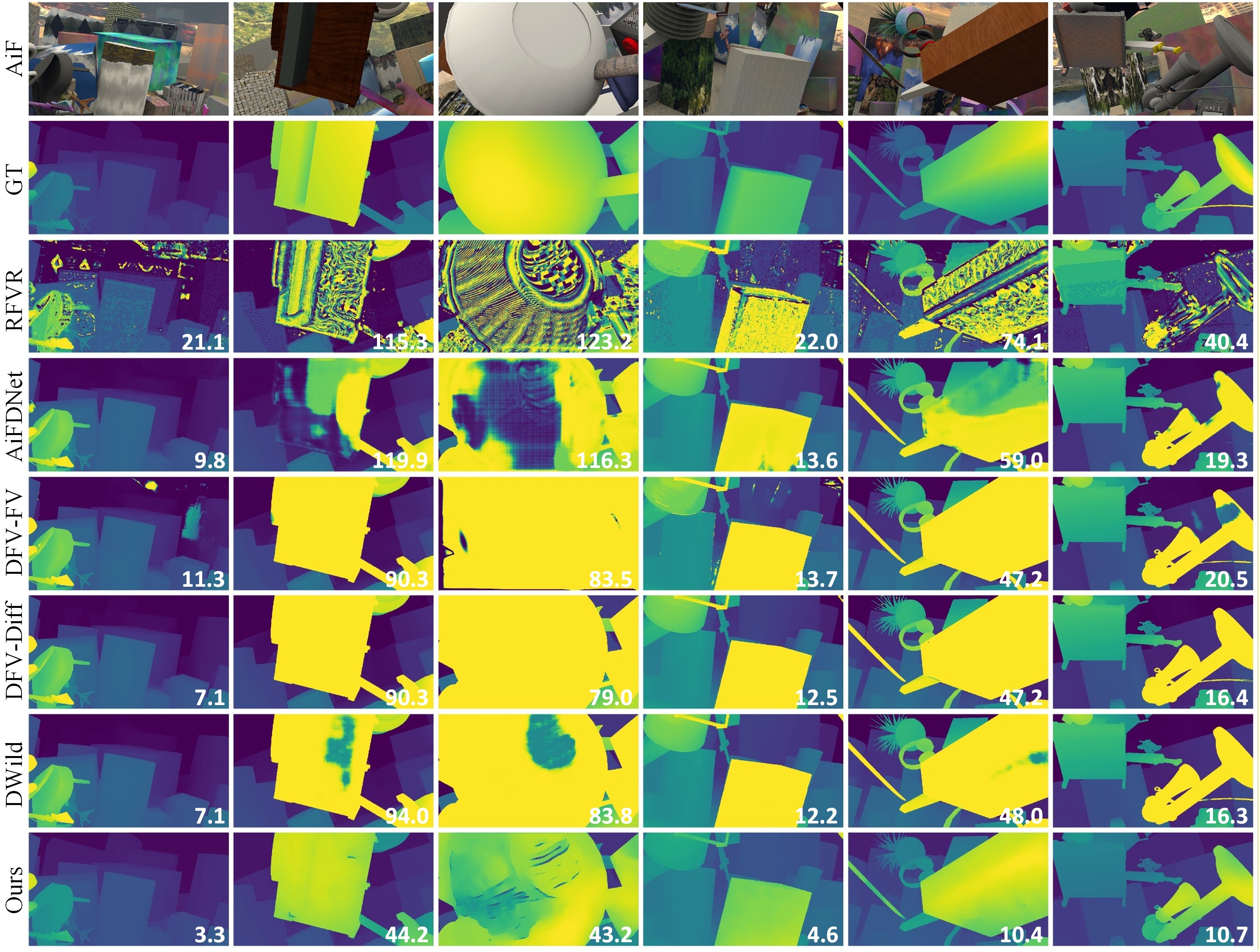}
        \caption{Qualitative comparison on the \textit{FT} dataset. The first row shows an All-in-Focus (AiF) image of the focal stack, the second shows the ground-truth (GT) map, while the remaining rows display the output maps produced by various methods. Numbers on each map denote the corresponding RMS error with respect to the GT.}

	\label{FT_Fig}
\end{figure*}

For \textit{FT} and \textit{Middlebury}, we used the official implementation of RFVR provided by the authors. Pretrained weights for AiFDNet and DWild were directly obtained from their respective official repositories. Since the authors did not release pretrained weights for DFV-FV and DFV-Diff for \textit{FT} dataset, we trained these models ourselves. For the \textit{FoD} dataset, the same protocol was followed: RFVR was executed using the official implementation, and pretrained weights for AiFDNet, DFV-FV, and DFV-Diff were obtained from their respective repositories. On the \textit{Mobile} dataset, the results for DDFF, DefocusNet, DFV-FV, and DFV-Diff were taken from \citep{yang2022deep}, while the results for DDFS were obtained from \citep{fujimura2024deep}.

\subsubsection{Comparison on the \textit{FT} Dataset}

We first evaluate our method on the \textit{FT} dataset, which is the largest among those considered, comprising 1000 training focal stacks and 100 testing focal stacks. For this comparison, except for RFVR, all models were exclusively trained on this dataset.

\begin{figure}[h]
        \centering
        \includegraphics[width=\columnwidth,keepaspectratio]{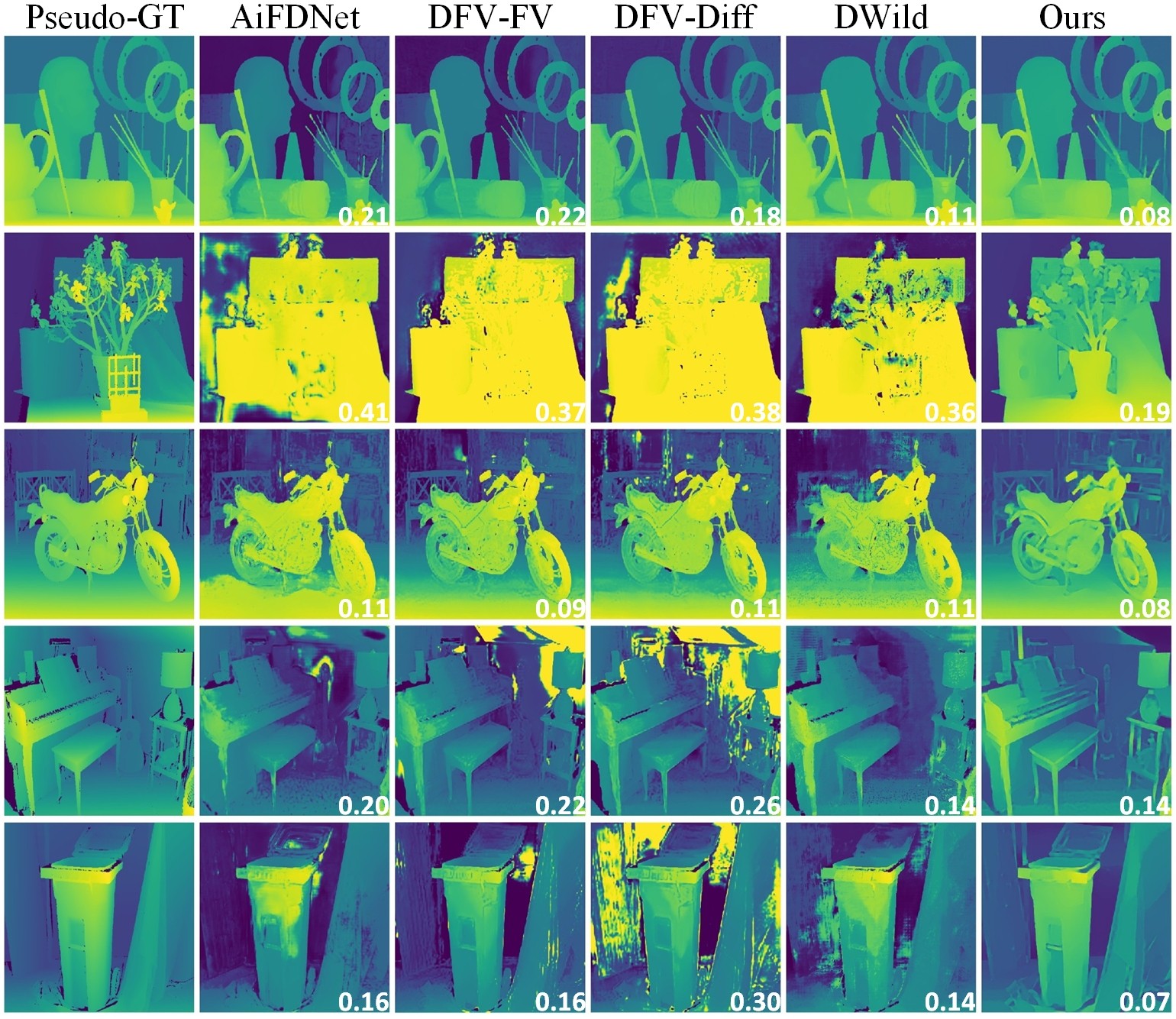}
        \caption{Qualitative generalization comparison on the real-world \textit{Middlebury} dataset. The first column shows pseudo-GT map taken from a structured light setup while the remaining columns display the output maps produced by various methods. Numbers below each map denote the corresponding normalized RMS error with respect to the pseudo-GT.}
	\label{MB_Fig}
\end{figure}

A detailed quantitative comparison is presented in \Cref{FT_Quant}. Note that the GT output maps span from 10 to 100 units, which accounts for the relatively high error values (e.g., MSE) across all methods. In the table, values highlighted in bold indicate the best performance. As the results show, our proposed method outperforms state-of-the-art approaches in 10 out of 12 metrics.

A qualitative comparison is also provided in \Cref{FT_Fig}. Among the evaluated methods, RFVR struggles with background noise and poorly defined object edges due to its strong dependence on the number of images in the focal stack, which limits its dynamic range and reduces its effectiveness on scenes with diverse depths. In contrast, the other five learning-based methods yield better output maps. As illustrated in \Cref{FT_Fig}, our proposed method produces the least noise, particularly in the second and fifth columns. In addition, our approach effectively handles complex internal patterns within objects; for example, in the fifth column a brown object contains intricate internal patterns, and our method accurately estimates its distance without being misled by the texture, an area where several comparative methods fall short. The superiority of our method is further reflected in the per-map Root Mean Squared (RMS) error values shown in the figure, where it consistently achieves the lowest error among all methods.

\begin{table}[htbp]
  \centering
  \caption{Quantitative comparison of depth maps on the test set of the \textit{FoD} dataset. Bold indicates the best performance. }
     \footnotesize
    \begin{tabular}{cccccc}
    \toprule
    Method & MAE   & MSE   & RMS   & logRMS & CORR \\
    \midrule
    RFVR  & 0.527 & 0.505 & 0.708 & 1.201 & 0.199 \\
    AiFDNet & \textbf{0.071} & 0.033 & 0.156 & \textbf{0.213} & 0.862 \\
    DFV-FV & 0.078 & 0.035 & 0.160 & 0.224 & 0.860 \\
    DFV-Diff & 0.077 & 0.037 & 0.165 & 0.225 & 0.859 \\
    Ours  & 0.074 & \textbf{0.030} & \textbf{0.151} & 0.231 & \textbf{0.864} \\
    \bottomrule
    \end{tabular}%
 \label{FoD_Quant}%
\end{table}%

\begin{figure}[hbtp]
        \centering
        \includegraphics[width=\columnwidth,keepaspectratio]{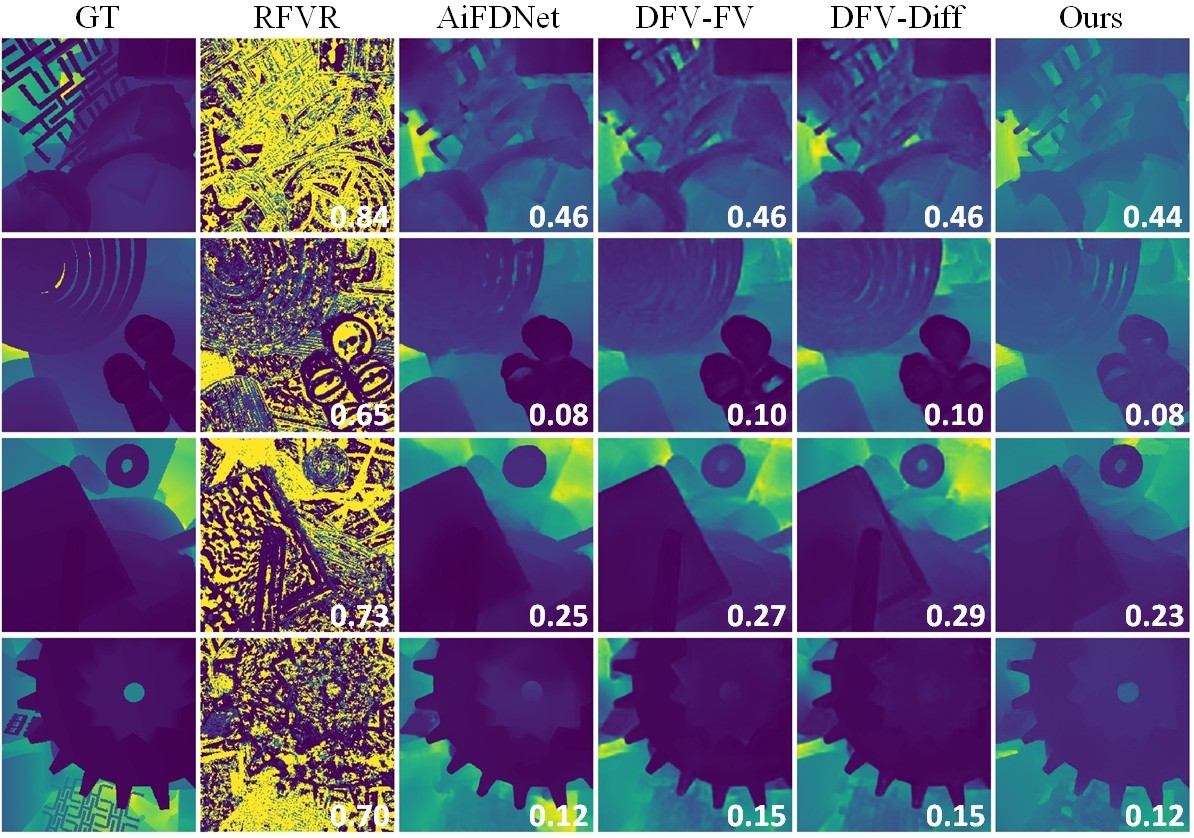}
        \caption{Qualitative comparison of depth maps on the test set of the \textit{FoD} dataset. First column shows GT depth, while remaining columns represent the method employed, while rows denote a different focal stack on which depths are computed. Numbers on each depth map denote the RMS error with respect to the GT.}

	\label{FoD_Fig}
\end{figure}

\begin{figure*}[hbtp]
        \centering
        \includegraphics[width=5.1in,keepaspectratio]{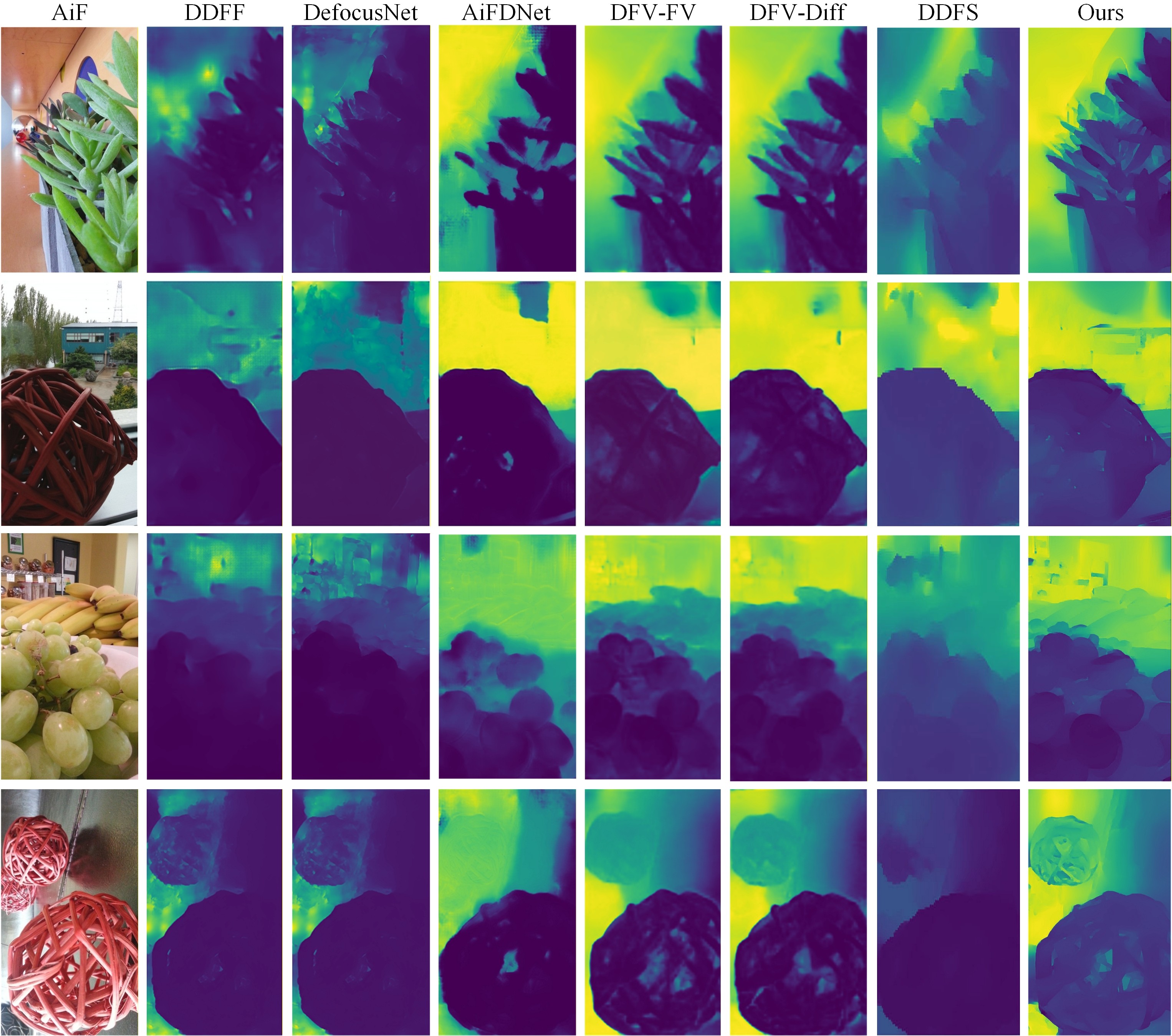}
        \caption{Qualitative generalization comparison on the real-world \textit{Mobile} dataset. The first column shows an All-in-Focus (AiF) image of the focal stack, while the remaining columns display the output maps produced by various methods.}

	\label{Mob_Qual}
\end{figure*}

\subsubsection{Generalization on the \textit{Middlebury} Dataset}

Our second comparative analysis evaluates the generalization performance of various learning based methods on the real-world \textit{Middlebury} dataset. For this evaluation, all models were trained exclusively on the \textit{FT} dataset. Although \textit{Middlebury} does not provide a predefined ground truth (GT) map, a pseudo-GT obtained using a structured light setup is available for reference. Qualitative results are shown in \Cref{MB_Fig}, where each output map is annotated with its normalized RMS error with respect to the pseudo-GT. The second row contains the most challenging scene, where all methods struggle to capture intricate object details. In this case, our method produces sharper edges and better object delineation, even though distinguishing between near and far objects remains difficult for all methods. In the fourth row, both DWild and our method perform better than the other approaches. In the fifth row, our method demonstrates reduced noise and improved edge clarity when reconstructing the dustbin. Overall, our method achieves the lowest per map RMS across the presented scenes, supporting the observed qualitative improvements.

\subsubsection{Comparison on the \textit{FoD} Dataset}

Our third comparative analysis evaluates performance on the synthetic \textit{FoD} dataset, which consists of focal stacks with only five images each. Quantitative results are presented in \Cref{FoD_Quant}. As indicated in the table, our proposed method ranks among the best in three out of five metrics.

A qualitative comparison is shown in \Cref{FoD_Fig}. RFVR struggles considerably because its depth estimation heavily depends on the number of images in the focal stack. In contrast, all supervised methods perform relatively better. Among them, our method achieves the best edge delineation, which is particularly evident in the third row where it most accurately captures the hole near the top-right corner. Performance gains are also reflected in the RMS error across the depth maps, where our method consistently achieves the lowest values for all focal stacks. In the last row, both AiFDNet and our method deliver the best results.

\begin{table*}[htbp]
    \centering
    \caption{Parameter count and average inference time (per focal stack) on the FoD dataset. 
    For our method, we report the time for DDL separately, along with the total time for different iteration settings.}
    \label{time}
    \resizebox{0.7\textwidth}{!}{%
    \begin{tabular}{lcccccccc}
        \toprule
        & AiFDNet & DFV-FV & DFV-Diff & \multicolumn{4}{c}{Ours} \\
        \cmidrule(lr){5-8}
        & & & & DDL & 1 it & 2 it & 4 it \\
        \midrule
        Params (M)    & 16.53 & 19.50 & 19.50 & 0.00 & 10.04 & 10.04 & 10.04 \\

        Mean Time (s) & 0.0272 & 0.0169 & 0.0179 & 0.0029 & 0.0231 & 0.0255 & 0.0349 \\
        \bottomrule
    \end{tabular}
    }
\end{table*}

\subsubsection{Generalization on the \textit{Mobile} Dataset}

Our final analysis is conducted on the \textit{Mobile} dataset as shown in \Cref{Mob_Qual}. For this evaluation, we developed a general model initially trained on the \textit{FT} dataset and subsequently fine-tuned on the \textit{FoD} and \textit{HCI} datasets. All comparative methods in this analysis are supervised, learning-based techniques. Our method shows consistent improvements across the four focal stacks. It is able to better separate background and foreground regions. In the first row, the edge boundaries between the foreground and the leaves’ background are more clearly delineated. In the first and third rows, our method also preserves edges more effectively. The third row further demonstrates its ability to capture fine details such as the delicate structures of the fruits in the scene. At the same time, similar to the other methods, our approach still encounters difficulties with complex patterned objects. In the last row, the balls with internal gaps remain challenging, and the predicted maps are not entirely precise.

\subsubsection{Runtime Analysis}

Table \ref{time} reports parameter counts and average inference times on the FoD test set, which contains five images per focal stack, each with a resolution of $256 \times 256$. In our method, the DDL module used to construct the focus volume is a parameter-free operator (0 parameters) and is very cheap to compute (0.0029 s per stack). The learnable part of the network contains 10.04M parameters and dominates the runtime. Our experiments show that the model produces strong results even with just 2 to 4 iterations. For the full pipeline, inference takes 0.0231 s per stack with a single iteration, 0.0255 s with two iterations, and 0.0349 s with four iterations.

\section{Conclusion}\label{sec5}

In this work, we presented a hybrid framework for shape-from-focus (SFF) that leverages the strengths of traditional and deep learning based SFF systems. Our approach begins by generating multi-scale focus volumes using handcrafted Directional Dilated Laplacian (DDL) kernels, which robustly capture long-range and directional focus variations. Building upon these robust focus measures, we introduced a lightweight, multi-scale GRU-based depth extraction module. By iteratively refining an initial zero-valued depth map, our recurrent network effectively integrates both local focus cues and global context features extracted from the mean image. Extensive experiments on synthetic and real-world datasets validate the superiority of our method over existing state-of-the-art approaches. Our framework not only achieves enhanced depth accuracy and spatial coherence but also demonstrates impressive generalization capabilities across diverse focal conditions. Future research will explore further optimization of the recurrent architecture and its extension to dynamic scenes and real-time applications, opening new avenues for robust depth estimation in complex environments.

\backmatter

\section*{Acknowledgements}
This work was supported by the Basic Research Program through NRF grant funded by the Korean government (MSIT: Ministry of Science and ICT) (2022R1F1A1071452)

\begin{appendices}

\end{appendices}

\bibliographystyle{sn-apacite}

\end{document}